\documentclass[reprint,aps,pre,floatfix]{revtex4-2}

\usepackage{amsmath,amssymb}
\usepackage{graphicx}
\graphicspath{{./}{../supplement/}}
\usepackage{algorithm}
\usepackage{algpseudocode}
\usepackage{booktabs}
\usepackage{hyperref}
\usepackage{placeins}

\begin{document}

\title{Variational Physics-Informed Ansatz for Reconstructing Hidden Interaction Networks from Steady States }

\author{Kaiming Luo}
\email{kmluo24@m.fudan.edu.cn}
\affiliation{School of Information Science and Technology, Fudan University, Shanghai 200438, China}

\date{September 10, 2026}
\begin{abstract}
Inferring interaction structure from steady-state observations is a central inverse problem when transient trajectories are unavailable. Here we formulate this problem as simultaneous compatibility of a single interaction operator with equilibrium constraints generated by heterogeneous perturbations. We introduce a variational physics-informed ansatz that represents the unknown operator as a trainable object and minimizes the resulting steady-state residuals across experiments. In the affine-interaction setting, the stacked equilibrium equations yield explicit finite-sample identifiability conditions: unique recovery is controlled by the rank of the compatibility matrix after elimination of experiment-wise gauge freedom. Synthetic benchmarks on pairwise, directed, weighted, empirical-topology, and selected higher-order systems illustrate this identifiability picture and show how additional heterogeneous steady states improve structural discrimination under the stated assumptions. The results clarify a concrete steady-state reconstruction regime in which equilibrium observations alone can determine hidden interaction operators when the governing dynamics are known and node-level equilibria are fully observed.
\end{abstract}

\maketitle

\noindent\textbf{Many complex systems reveal themselves not through complete trajectories but through the states they reach after repeated perturbations. This creates a basic inverse problem for nonlinear network dynamics: can hidden interactions be recovered when only the final equilibria are observed? This work shows that, when the governing equations are known and perturbations generate sufficiently diverse steady states, the answer is controlled by compatibility across experiments. A variational physics-informed ansatz enforces this compatibility directly, while finite-sample rank conditions explain when the hidden interaction operator becomes identifiable. The framework connects steady-state measurements, nonlinear dynamics, and network reconstruction without requiring time-resolved trajectories.}

\section{Introduction}

Inferring hidden interaction structure from observations is a central inverse problem in complex systems\cite{arenas2008synchronization,strogatz2001exploring}. In neural, ecological, biochemical, and engineered networks, collective behavior depends on both the intrinsic node dynamics and the interaction architecture that couples the degrees of freedom\cite{barabasi1999emergence,watts1998collective,newman2003structure,luo2024remote}. Recovering this architecture from measured states is therefore a basic problem in nonequilibrium statistical physics, network dynamics, and dynamical-systems inference.

A common experimental setting is repeated perturbation followed by relaxation to a stable equilibrium or fixed point. In many settings, perturbations can be imposed reproducibly, whereas transient trajectories are difficult to sample densely or reliably. What is then available is a collection of heterogeneous steady states generated under different interventions. Examples include gene-expression measurements, ecological equilibria, and engineered operating points. In such cases, the accessible equilibrium observations define the reconstruction problem.

This regime differs qualitatively from trajectory-based reconstruction. If transient dynamics are observed with sufficient temporal resolution, one may estimate derivatives, local Jacobians, or response kernels. In steady-state-only experiments, the information about the unknown coupling operator is distributed across multiple equilibria. The reconstruction problem is therefore to determine whether a single interaction operator can satisfy the steady-state equations for all experiments simultaneously. This is an across-experiment compatibility problem. Its difficulty is controlled by identifiability: if the measured equilibria are not sufficiently heterogeneous, distinct candidate structures may remain compatible with the same data.

Although many reconstruction methods perform well in the regimes for which they were designed\cite{brugere2018network}, several difficulties remain when the data are noisy, the dynamics are nonlinear, the interactions are higher order, or only steady states are accessible. Classical time-series methods\citep{timme2007revealing,yuan2011robust,shandilya2011inferring} rely on densely sampled trajectories to estimate derivatives or local regression models, and their performance deteriorates under sparse sampling, latent variables, or measurement noise. Extensions to nonlinear and multi-body couplings\citep{malizia2024reconstructing} address higher-order interactions, which can qualitatively reorganize collective dynamics\cite{luo2026shear}, but still require temporally resolved data. Equilibrium-state, regression, and model-selection approaches\citep{sontag2008network,yu2008driving,nitzan2017revealing,peixoto2019bayesian,Peixoto2025Network} provide complementary inference principles, but often depend on linearization, tractable analytical response formulas, or strong structural priors. Scalability is an additional limitation: many representative schemes require dense regression, matrix inversion, or other operations with at least cubic cost and are therefore difficult to extend to dense, weighted, or higher-order networks\citep{napoletani2008reconstructing,wang2011predicting,schreiber2000measuring,tajima2015untangling,brunton2016discovering}.

Recent progress in constraint-based and equation-aware learning suggests a way forward. Physics-informed approaches incorporate known dynamical constraints directly into differentiable parameterizations\citep{lagaris1998artificial,raissi2019physics,karniadakis2021physics,luo2026controlling}, with related equation-constrained optimization used in finite-time thermodynamic control\cite{luo2026erasure}. Interpretable model-discovery and Bayesian formulations show how structural priors can improve robustness and data efficiency\citep{brunton2016discovering,peixoto2019bayesian,Peixoto2025Network}. These ideas motivate a compatibility-based inverse formulation in which the interaction operator is the primary unknown and repeated equilibria supply the observations.

Related reconstruction problems use different observations and priors. Causal statistical inference can reconstruct missing networks under adversarial interventions\cite{xue2019adversarial}, while multifractal graph models characterize generating measures\cite{yang2020multifractal} and infer hidden network-generating rules from partial observations\cite{yang2021hidden}. These approaches target latent topology or its statistical generator. Here the unknown is a dynamical interaction operator constrained by fully observed equilibria and known interventions.

Geometric topology and side information can identify functional communities in incomplete biological networks\cite{sia2022communities}; simplicial representation learning captures higher-order structure in relational data\cite{yang2022simplicial}. Multifractal statistics of individual-neuron spiking can also discriminate network architecture and function\cite{yang2025spiking}. These approaches infer statistical or functional organization from partial or temporal observations; the present problem concerns interaction coefficients constrained by prescribed dynamical equations and fully observed equilibria.

The present work addresses a controlled steady-state reconstruction regime. We assume that the governing functional form is known, perturbations are externally specified, and node-level equilibria are fully observed after relaxation. We characterize and test the regime in which heterogeneous equilibria provide enough independent compatibility constraints to identify the interaction operator.

In this work, we introduce a steady-state variational reconstruction scheme, termed the Variational Physics-Informed Ansatz (VPIA), for inferring interaction operators from heterogeneous steady-state measurements under known dynamics. VPIA represents the interaction operator as a trainable variational object and infers it by enforcing simultaneous compatibility with the governing steady-state equations across perturbations. In this sense ``physics-informed'' means that candidate operators are evaluated by residuals of the equilibrium equations. The same formulation accommodates admissibility mappings, directed and weighted pairwise structures, residual sampling, and selected higher-order operators without relying on temporal trajectories or derivative estimation.

\section{Steady-State Variational Reconstruction Framework}

\begin{figure}
    \centering
    \includegraphics[width=\linewidth]{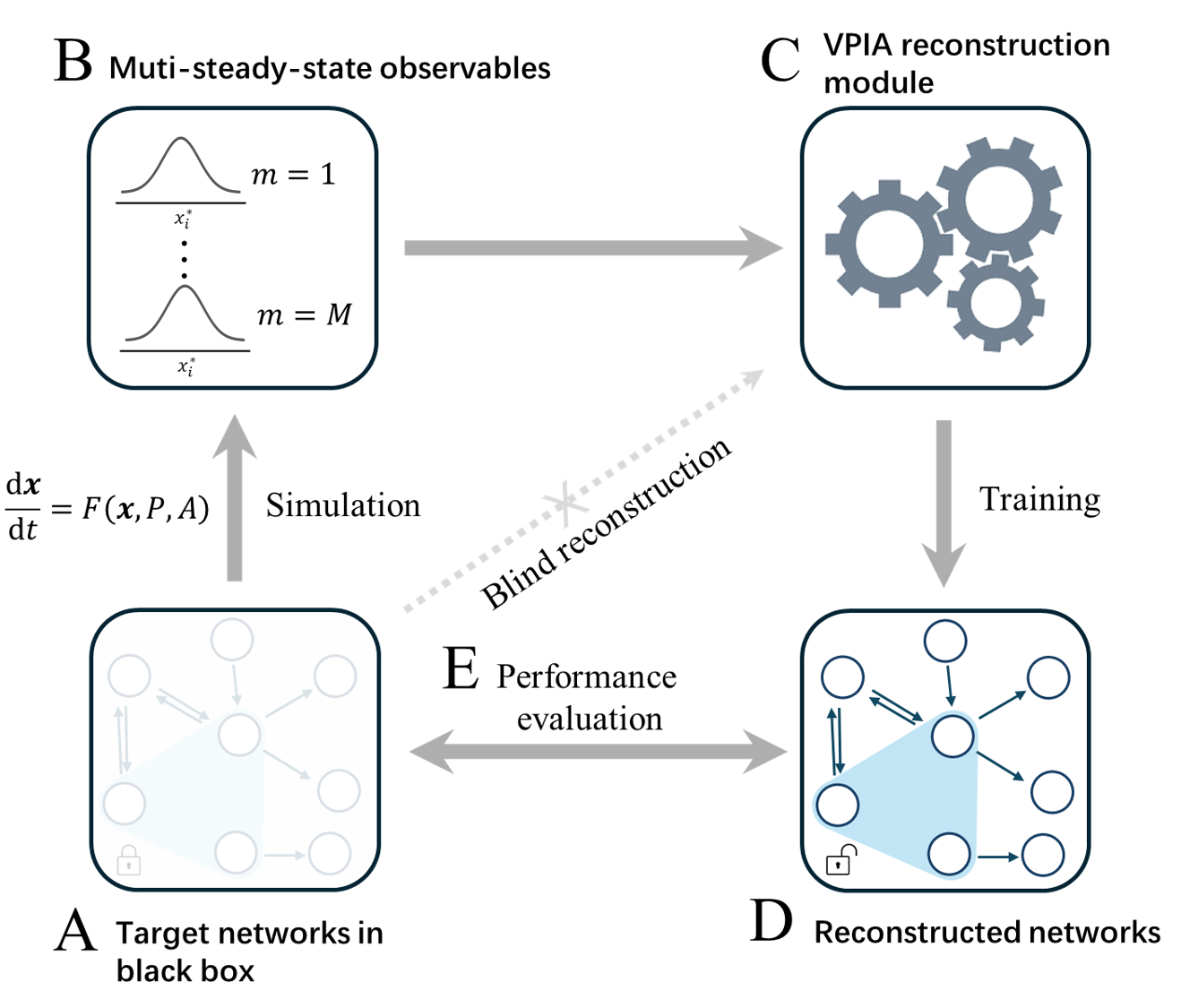}
    \caption{\textbf{Overview of the VPIA steady-state reconstruction framework.}
    \textbf{(A)} Representative planted interaction structure used only for validation.
    The framework is tested on undirected, directed, weighted, and selected higher-order interaction patterns.
    The light-blue shaded region indicates a representative 2-simplex.
    \textbf{(B)} Collection of heterogeneous steady-state observations generated under different intrinsic parameters or driving conditions.
    Each trial $m$ relaxes to a stable configuration $x^{*(m)}$, which is used for reconstruction.
    \textbf{(C)} Variational representation and equilibrium-constrained optimization.
    The reconstruction module receives the observed steady states, known controls, and governing equations, without access to the true interaction structure.
    Each candidate interaction is assigned a trainable parameter and mapped to a continuous admissibility variable through a differentiable ansatz.
    \textbf{(D)} Reconstructed interaction operator $\widehat{A}$ obtained by minimizing the steady-state residuals across experiments.
    \textbf{(E)} Quantitative evaluation of $\widehat{A}$ against the ground-truth structure $A$, with AUC used as the primary reconstruction metric.}
    \label{fig:overview}
\end{figure}

We formulate network reconstruction from repeated steady-state observations as a constrained inverse problem. The setting is illustrated in Fig.~\ref{fig:overview}. Throughout this section, ``physics-informed'' means that candidate interaction operators are evaluated by their consistency with the governing steady-state equations under multiple perturbation experiments.

We consider an $N$-node dynamical system observed under $M$ controlled perturbation experiments. Let $x_i \in \mathbb{R}^{q_i}$ denote the state of node $i$, let $p^{(m)}$ denote the externally imposed condition in experiment $m$, and let $\{A^{(d)}\}_{d\ge 1}$ denote the unknown interaction objects to be reconstructed. The dynamics are written in the general form
\begin{equation}
    \dot{x}_i = F_i\!\left(x;\, p^{(m)}, \{A^{(d)}\}_{d\ge 1}\right),
\end{equation}
where $A^{(1)}$ represents pairwise interactions and higher-order tensors $A^{(2)},A^{(3)},\ldots$ are introduced only when selected simplex- or hypergraph-like couplings are considered. For notational simplicity, the main construction is written in scalar form when convenient, but the formulation extends directly to vector-valued node states.

In each experiment, the system is initialized, evolved, and recorded only after convergence to a stable steady state. Numerically, convergence is declared once $\|\dot{x}\|_2 < \varepsilon_{\mathrm{conv}}$, and the measured output of experiment $m$ is the equilibrium configuration $x^{*(m)}$. The reconstruction problem is then to infer a single interaction operator that is compatible with all observed equilibria simultaneously.

The informativeness of the data depends on the heterogeneity of the measured steady states. If the perturbations generate nearly synchronized or otherwise weakly differentiated equilibria, the resulting steady-state constraints may contain little structural information and can become effectively rank-deficient. We therefore retain only steady states satisfying
\begin{equation}
    D\!\left(x^{*(m)}\right) > \epsilon_{\mathrm{sync}},
\end{equation}
where $D$ is a dispersion measure, such as the state variance or the mean pairwise distance, and $\epsilon_{\mathrm{sync}}$ is a prescribed heterogeneity threshold. Such a filter does not alter the governing equations and is optional. The stability checks in Sec.~S1 of the Supplementary Material use explicit stability and convergence checks without an additional dispersion cutoff; information quality is assessed through reconstruction and rank diagnostics.

The unknown interaction operator is represented variationally. For unweighted pairwise interactions, each candidate edge is assigned a trainable parameter $\theta_{ij}$ and mapped to a soft adjacency variable through
\begin{equation}
    \hat{A}_{ij} = \sigma(k\theta_{ij}) = \frac{1}{1+\exp(-k\theta_{ij})},
\end{equation}
where $k$ controls the sharpness of the continuous-to-binary transition. Undirected networks are enforced by symmetry, directed networks use independent ordered parameters, and weighted networks retain continuous outputs without binarization. Selected higher-order interactions are parameterized analogously, for example,
\begin{equation}
    \hat{A}^{(2)}_{ijk} = \sigma(k\theta_{ijk}), \qquad
    \hat{A}^{(3)}_{ijkl} = \sigma(k\theta_{ijkl}),
\end{equation}
which yields a unified differentiable representation across the interaction classes considered here.

Reconstruction is driven directly by the equilibrium constraints. Given a candidate interaction operator $\hat{A}$, each observed steady state $x^{*(m)}$ defines node-wise residuals
\begin{equation}
    R_i^{(m)} =
    F_i\!\left(x^{*(m)},\, p^{(m)},\, \hat{A}\right).
\end{equation}
The full compatibility objective is
\begin{equation}
    \mathcal{L}
    =
    \frac{1}{MN}\sum_{m=1}^{M}\sum_{i=1}^{N}
    \left[R_i^{(m)}\right]^2.
\end{equation}
Minimizing $\mathcal{L}$ enforces simultaneous consistency of a single interaction operator with all retained steady-state experiments. In contrast to derivative-based local regression, the present objective is defined globally at the level of the governing steady-state equations and couples all nodes through the candidate operator. The reconstructed interaction operator $\widehat{A}$ is obtained from the variational parameters that minimize this loss.

\subsection{Finite-sample identifiability in the affine-interaction setting}

This subsection develops finite-sample identifiability for dynamics that depend affinely on the unknown interaction coefficients.

\smallskip
\noindent\textbf{Theorem 1 (finite-sample identifiability in the affine setting).}
Assume that (i) the governing dynamics are known, (ii) the steady-state equations depend affinely on the unknown interaction coefficients, (iii) node-level steady states are fully observed, (iv) experiment-wise scalar gauges, if present, are eliminated, and (v) the independent unknown coefficients are vectorized into a vector $\mathbf a\in\mathbb{R}^{E}$. Then, after stacking $M$ experiments, the reduced equilibrium constraints take the form $H\mathbf a=\mathbf b$. For unrestricted continuous coefficients, the interaction operator is uniquely identifiable if and only if $\mathrm{rank}(H)=E$.

In scalar notation, the affine dynamics can be written as
\begin{equation}
    \dot{x}_i
    =
    H_i(x,\lambda)
    +
    \sum_j A_{ij}\,\Phi_{ij}(x,\lambda),
    \label{eq:affine_dependence_main}
\end{equation}
where $\lambda$ collects the experiment-dependent controls and $\Phi_{ij}$ is known once $x$ and $\lambda$ are specified. For experiment $m$ with steady state $x^{(m)}$, the equilibrium equations can be written as
\begin{equation}
    0
    =
    H_i^{(m)}
    +
    \sum_j A_{ij}\,\Phi_{ij}^{(m)}
    -
    c^{(m)},
\end{equation}
where $c^{(m)}$ is an experiment-wise scalar gauge, when required by the observation model. For phase-locked oscillators this is the common rotation frequency. If no unknown gauge is present, the full equations should instead be retained; their independent row count replaces $N-1$ below. Subtracting a reference equation in that case remains a valid reduced construction but can discard information. Eliminating $c^{(m)}$ against node $N$ yields $(N-1)$ linear constraints per steady state,
\begin{equation}
    \begin{aligned}
    &\bigl(H_i^{(m)}-H_N^{(m)}\bigr)
    +
    \sum_j
    \bigl(
    A_{ij}\Phi_{ij}^{(m)}-A_{Nj}\Phi_{Nj}^{(m)}
    \bigr)
    =
    0,\\
    &\hspace{7em} i=1,\dots,N-1.
    \end{aligned}
    \label{eq:ss_reduced_main}
\end{equation}
After vectorizing the independent interaction coefficients into $\mathbf a$, the stacked constraints from $M$ experiments take the linear form
\begin{equation}
    H\mathbf a = \mathbf b,
    \qquad
    H\in\mathbb{R}^{(N-1)M\times E}.
    \label{eq:stacked_identifiability_main}
\end{equation}
Unrestricted continuous-coefficient recovery is equivalent to $\mathrm{rank}(H)=E$. Indeed, if there exists a nonzero vector $v$ such that $Hv=0$, then both $\mathbf a$ and $\mathbf a+v$ satisfy the same stacked steady-state equations, so the interaction coefficients cannot be distinguished. Full column rank is therefore exactly the condition that removes nontrivial null directions. Since
\begin{equation}
    \mathrm{rank}(H)\le \min\{(N-1)M,E\},
    \label{eq:rank_counting_main}
\end{equation}
a necessary row-count condition for the reduced system is $(N-1)M\ge E$. Define the counting lower bound
\begin{equation}
 M_{\mathrm{count}}=\left\lceil\frac{E}{N-1}\right\rceil.
 \label{eq:mstate_gen_main}
\end{equation}
For the all-pairs undirected and directed candidate sets, respectively, this gives
\begin{equation}
 M_{\mathrm{count}}^{\mathrm{undir}}=\left\lceil N/2\right\rceil,
 \qquad M_{\mathrm{count}}^{\mathrm{dir}}=N.
 \label{eq:count_pairwise}
\end{equation}
For one independent coefficient per unordered $n$-simplex, the corresponding expression is
\begin{equation}
 M_{\mathrm{count}}^{(n)}=\left\lceil\frac{\binom{N}{n+1}}{N-1}\right\rceil.
 \label{eq:mstate_higher_gen_main}
\end{equation}
Here $E$ counts candidate coefficients, not the number of nonzero coefficients in the unknown network. Equations~\eqref{eq:mstate_gen_main}--\eqref{eq:mstate_higher_gen_main} are necessary counting bounds for this reduced continuous-coefficient problem, not generic sufficient thresholds.

A conditional generic statement is possible. Fix the model, candidate interactions, number of experiments, and a connected open domain of admissible controls on a single analytic equilibrium branch. If the resulting matrix $H$ depends analytically on these controls and one admissible configuration has a nonzero $E\times E$ minor, then that minor vanishes only on a measure-zero subset of this domain. Full column rank is thus generic on this branch. Analytic dependence follows locally from analytic dynamics and the implicit-function theorem when the equilibrium Jacobian, after any gauge reduction, is nonsingular. Existence of the required minor and accessibility of the equilibrium branch are additional conditions beyond row counting.

Scalar diffusive coupling supplies a concrete obstruction. For an undirected candidate edge $(i,j)$, let $d_{ij}$ be its oriented incidence vector and let $p_i=(x_i^{(1)},\ldots,x_i^{(M)})$. The corresponding stacked column is, up to orientation and row ordering, $d_{ij}\otimes(p_j-p_i)$. Translations and infinitesimal rotations of these points give null directions of the transpose. For generic point configurations with $M\le N-1$, this implies
\begin{equation}
 \operatorname{rank}(H)\le
 \min\!\left\{E,MN-\frac{M(M+1)}{2}\right\}.
 \label{eq:diffusive_rank}
\end{equation}
Indeed, a displacement $v_i=c+\Omega p_i$, with $\Omega^\top=-\Omega$, satisfies $(p_j-p_i)^\top(v_j-v_i)=0$. The independent translations and rotations have dimension $M(M+1)/2$ for an affinely spanning configuration. At $N=10$, $M=5$, the bound is $35$, below the $45$ unknown all-pairs weights despite meeting the row count. A numerical rank check against Eq.~\eqref{eq:diffusive_rank} is included in Sec.~S4 of the Supplementary Material. This obstruction concerns scalar difference coupling; other interaction functions require their own rank analysis.

If coefficients lie in a constrained admissible set $\mathcal A$, uniqueness at $\mathbf a^\star$ instead requires $\ker H\cap(\mathcal A-\mathbf a^\star)=\{0\}$. Binary or sparsity constraints can therefore permit support recovery without unrestricted continuous identifiability. The empirical support-recovery counts below concern this constrained setting and depend on the experimental protocol.

\subsection{Relation to direct linear residual solvers}

In the exactly affine case, the stacked compatibility equations define a linear inverse problem. Direct solvers such as least squares, ridge regularization, LASSO, or other sparsity-regularized regressions are therefore natural baselines and may be preferable for small, clean affine systems. VPIA should not be interpreted as replacing these direct solvers in that limit. Rather, the affine reduction provides a transparent identifiability criterion, while the variational formulation supplies a common implementation for constrained admissibility maps, continuous edge scores, directed and weighted operators, selected higher-order tensors, residual sampling, and differentiable extensions where explicitly assembling or solving the full linear system is inconvenient.

The numerical comparisons therefore include solvers supplied with the same equilibrium equations and controls, as well as association-based and response-based adaptations. They separate the information supplied by equilibrium compatibility from the choice of parameterization and optimizer.

\subsection{Residual sampling and Fisher-preconditioned optimization}

For large systems, evaluating all $MN$ residuals at each iteration is unnecessary. We therefore use residual sampling. At optimization step $t$, a subset $S_t \subset \{1,\ldots,N\}$ with $|S_t| = s \ll N$ is drawn uniformly and evaluated across the $M$ retained steady states. The number of sampled node-state constraints in one optimization step is therefore
\begin{equation}
    r = Ms.
\end{equation}
The sampled objective
\begin{equation}
    \mathcal{L}_{\mathrm{RS}}
    =
    \frac{1}{Ms}\sum_{m=1}^{M}\sum_{i\in S_t}
    \left[R_i^{(m)}\right]^2
\end{equation}
is minimized instead. Under uniform sampling, $\mathcal{L}_{\mathrm{RS}}$ is an unbiased estimator of the full loss $\mathcal{L}$, while the pairwise residual-evaluation workload is $\mathcal{O}(Ms k_c)=\mathcal{O}(r k_c)$, where $k_c$ is the mean number of candidate neighbours accessed per sampled node. For all-pairs inference, $k_c=N-1$ regardless of true-network sparsity. Thus $r$ includes all sampled node-state residuals used in a single optimization step; the dependence on $M$ is contained in $r$ under the node-sampling implementation used here.

To improve optimization conditioning, we combine this sampled objective with a diagonal Gauss--Newton/Fisher-type preconditioner. Approximating the residual model locally by a Gaussian least-squares structure, we use the diagonal approximation below, where $\sigma^2$ denotes the effective residual variance used in the local least-squares approximation; an overall normalization can be absorbed into the step size. The diagonal need not improve every problem, as tested by the optimizer ablation below.
\begin{equation}
    F(\theta)\approx
    \mathrm{diag}\!\left[
    \frac{1}{\sigma^2}\sum_{m=1}^{M}\sum_{i\in S_t}
    \nabla_{\theta}R_i^{(m)}
    \nabla_{\theta}R_i^{(m)\top}
    \right],
\end{equation}
which gives the practical update
\begin{equation}
    \Delta\theta
    =
    -\eta\,
    \frac{\nabla_{\theta}\mathcal{L}_{\mathrm{RS}}}
    {\varepsilon_{\mathrm{reg}}+\mathrm{diag}(F(\theta))}.
\end{equation}
This update provides a diagonal curvature approximation while avoiding the construction and inversion of the full dense Fisher matrix. The associated per-iteration scaling is discussed in Sec.~S4 of the Supplementary Material.

\section{Results}

The steady-state variational reconstruction framework is evaluated along four directions: core reconstruction on pairwise networks, comparison with representative baselines, robustness under noise, and extension to benchmark empirical topologies and selected higher-order interactions. Throughout this section, reconstruction quality is quantified primarily by the area under the receiver operating characteristic curve (AUC)\cite{hanley1982meaning,fawcett2006introduction}, computed from the continuous predicted interaction scores before thresholding,
\begin{equation}
    \mathrm{AUC}
    =
    \int_0^1 \mathrm{TPR}\!\bigl(\mathrm{FPR}\bigr)\, d(\mathrm{FPR}).
\end{equation}
Because AUC depends on the ranking of inferred interaction scores rather than on a single decision threshold, it provides a threshold-independent measure of structural discrimination across networks with different sparsities. Fixed thresholds are used only for visualization or for reporting support recovery in selected examples.

\subsection{Core reconstruction behavior on pairwise networks}

\begin{figure*}[t]
\centering
\includegraphics[width=0.75\linewidth]{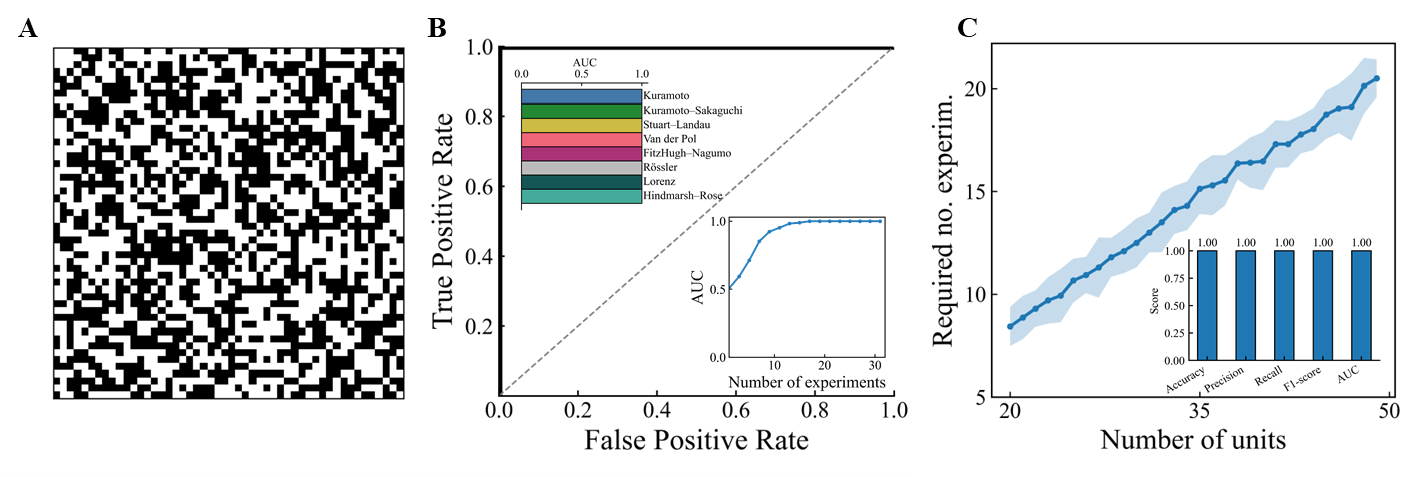}
\caption{\textbf{Reconstruction performance of the VPIA framework.}
(A) Ground-truth adjacency matrix, with black and white entries denoting the presence and absence of links.
(B) ROC curve for Kuramoto steady-state data. The upper-left inset reports AUC for the indicated dynamical models, and the lower-right inset shows AUC versus the number of experiments.
(C) Structural-recovery experiment counts versus network size, averaged over 20 realizations with standard-deviation bands; recovery requires AUC $\ge0.999$ and zero support errors at threshold $0.5$. The inset reports accuracy, precision, recall, F1 score, and AUC. No added noise is used.}
\label{fig:network_reconstruction}
\end{figure*}

We first examine pairwise reconstruction using the Kuramoto model\cite{kuramoto1984chemical} as the reference example. A representative ground-truth adjacency matrix is shown in Fig.~\ref{fig:network_reconstruction}A, where black and white entries indicate the presence and absence of pairwise links, respectively. This planted structure is used only for validation; the reconstruction procedure itself receives only the steady-state observations generated under repeated perturbations.

The phase dynamics are written as
\begin{equation}
    \dot{\phi}_i
    =
    \omega_i^{(m)}
    +
    K \sum_{j=1}^{N} A_{ij}\sin(\phi_j-\phi_i),
    \qquad i=1,\dots,N,
\end{equation}
where $\phi_i$ is the phase of node $i$, $\omega_i^{(m)}$ is the intrinsic frequency used in perturbation experiment $m$, $K$ is the coupling strength, and $A_{ij}$ is the adjacency matrix to be reconstructed. Each perturbation experiment produces a stable steady state
\(
\phi^{*(m)}
\)
satisfying
\begin{equation}
    0
    =
    \omega_i^{(m)}
    +
    K \sum_{j=1}^{N} A_{ij}\sin\!\bigl(\phi_j^{*(m)}-\phi_i^{*(m)}\bigr).
\end{equation}
These equilibrium equations define the compatibility constraints used by VPIA.

As additional heterogeneous steady states are incorporated, discrimination performance improves and approaches a saturated regime in Fig.~\ref{fig:network_reconstruction}B. Its upper-left inset includes Kuramoto--Sakaguchi, Stuart--Landau, Van der Pol, FitzHugh--Nagumo, R\"ossler, Lorenz, and Hindmarsh--Rose systems\cite{sakaguchi1986soluble,landau1944problem,stuart1960nonlinear,vanderpol1926lxxxviii,fitzhugh1961impulses,nagumo1962active,rossler1976equation,lorenz1963deterministic,hindmarsh1984model}. The equations, parameter distributions, stable operating regimes, and convergence checks for Fig.~\ref{fig:network_reconstruction} are detailed in Sec.~S1 of the Supplementary Material.

The experiment counts in Fig.~\ref{fig:network_reconstruction}C increase with network size and quantify thresholded binary-support recovery. Admissibility constraints can distinguish binary structures even when continuous coefficients remain nonunique; unrestricted continuous recovery requires the full-column-rank condition of Sec.~II.

For Fig.~\ref{fig:network_reconstruction}C, saturation is defined operationally by AUC $\ge0.999$, while exact support recovery requires zero errors at the fixed threshold $\hat A_{ij}>0.5$. The required experiment count is the first tested $M$ satisfying both conditions under the specified reconstruction protocol. These protocol-dependent counts are averaged over 20 realizations; further details are given in Sec.~S1 of the Supplementary Material. Additional pairwise tests on Erd\H{o}s--R\'enyi, scale-free, and small-world networks, as well as directed and weighted extensions, are retained in Sec.~S2 of the Supplementary Material as numerical demonstrations within their stated settings.

\subsection{Robustness under noise}

\begin{figure}
    \centering
    \includegraphics[width=0.85\linewidth]{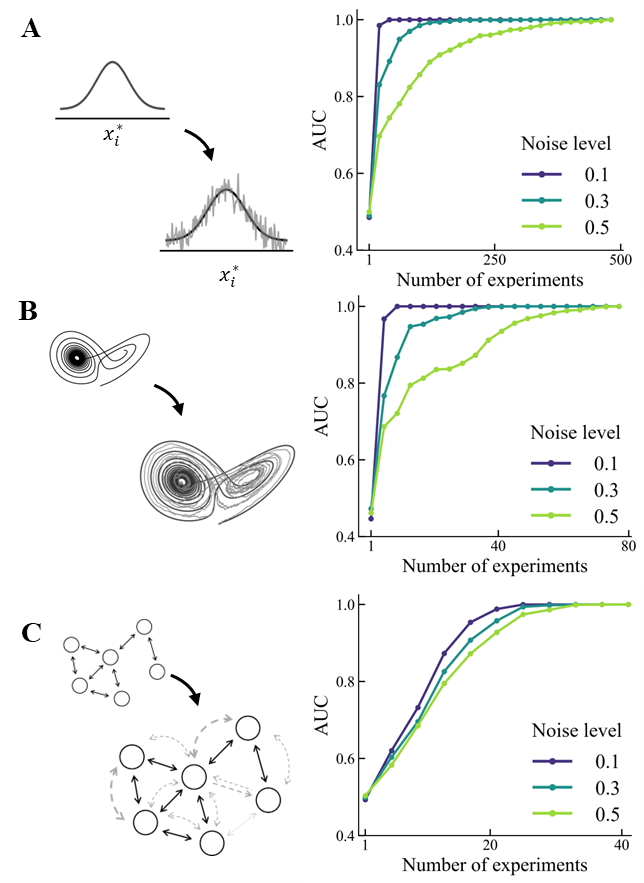}
    \caption{\textbf{Synthetic robustness tests under three classes of zero-mean noise.}
    (A) Observation noise, in which each measured steady state is contaminated by additive random fluctuations.
    (B) Dynamical noise, modeled as a stochastic term acting during the time evolution.
    (C) Structural noise, corresponding to perturbations of the underlying interaction coefficients.
    For each noise class, three representative amplitudes are shown in the right-hand panels.
    Across all cases, reconstruction accuracy improves as the number of heterogeneous steady-state experiments increases.
    For structural noise, increasing $M$ suppresses sampling fluctuations but does not remove the perturbation-induced accuracy floor relative to the clean underlying structure.
    All curves are averaged over 20 independent realizations; $M$ varies along the horizontal axis in each panel, $N$ is fixed within each panel, and the noise amplitudes are $0.1$, $0.3$, and $0.5$, as indicated in the legends.}
    \label{fig:Noise}
\end{figure}

Robustness is next examined under observation, dynamical, and structural noise. In all three settings, reconstruction improves systematically as the number of steady-state experiments increases in Fig.~\ref{fig:Noise}, indicating that cross-experiment compatibility remains informative under noisy conditions.

The three noise classes are defined as follows. Observation noise perturbs the measured equilibrium values according to
\begin{equation}
    \tilde{x}^{*(m)}_i
    =
    x^{*(m)}_i + \sigma_{\mathrm{obs}}\,\mathrm{MAD}\,\xi_i,
    \qquad \xi_i\sim\mathcal{N}(0,1),
\end{equation}
where $\mathrm{MAD}$ denotes the median absolute deviation used to normalize the noise amplitude. Dynamical noise acts during the relaxation process,
\begin{equation}
    \dot{x}_i
    =
    F_i\bigl(x;\,p^{(m)},A\bigr)
    + \sigma_{\mathrm{dyn}}\,\xi_i(t),
\end{equation}
where $\xi_i(t)$ is a zero-mean Gaussian white-noise process. Structural noise perturbs the interaction coefficients themselves,
\begin{equation}
    \tilde{A}_{ij}
    =
    A_{ij}+\sigma_{\mathrm{str}}\,\xi_{ij},
    \qquad \xi_{ij}\sim\mathcal{N}(0,1).
\end{equation}
In all three cases, the perturbations are zero-mean and are regenerated independently across realizations. This common parameterization permits a direct comparison between measurement uncertainty, process noise, and perturbations of the underlying structure within the same reconstruction protocol.

\smallskip
\noindent\textbf{Proposition 1 (small-noise perturbation of the affine estimator).}
Let the clean system satisfy $H\mathbf a^\star=\mathbf b$, with $H$ of full column rank. State measurement errors generally perturb both sides: $\tilde H=H+\Delta H$ and $\tilde{\mathbf b}=\mathbf b+\Delta\mathbf b$. The unregularized least-squares estimate obeys, to first order,
\begin{equation}
 \hat{\mathbf a}-\mathbf a^\star
 =H^\dagger(\Delta\mathbf b-\Delta H\mathbf a^\star)
 +O(\|\Delta\|^2).
 \label{eq:noise_corrected}
\end{equation}
For independent small, centered state errors and $M^{-1}H^\top H\to Q\succ0$, the first-order standard deviation is $O(\sigma_{\mathrm{obs}}/\sqrt M)$ at fixed parameter dimension, but the nonlinear errors-in-variables bias can remain $O(\sigma_{\mathrm{obs}}^2)$. For time averages of mixing fluctuations around a stable operating state, replace the state variance by its effective time-averaged value; stationary mean shifts must be treated separately. Sec.~S3 of the Supplementary Material states the assumptions and derivation. This local expansion applies to the unregularized affine estimator; constraints and finite optimization introduce additional effects.

\smallskip
\noindent\textbf{Proposition 2 (structural-noise floor).}
Structural noise is qualitatively different because the physical interaction operator is itself modified. If the perturbed coefficients are written as $\tilde{\mathbf a}=\mathbf a^\star+\boldsymbol\delta$ with $\mathbb{E}[\boldsymbol\delta]=0$ and $\mathrm{Cov}(\boldsymbol\delta)=O(\sigma_{\mathrm{str}}^2)I$, then, in the absence of measurement bias and optimization error, a consistent reconstruction targets the perturbed operator for each fixed realization, while the error relative to the clean structure separates into a sampling term and an irreducible structural term,
\begin{equation}
    \mathbb{E}\!\left[\|\hat{\mathbf a}-\mathbf a^\star\|_2^2\right]
    =
    O\!\left(\frac{1}{M}\right)
    +
    O(\sigma_{\mathrm{str}}^2).
\end{equation}
Increasing the number of steady states therefore suppresses the sampling-induced component of the error but does not remove the discrepancy between the clean and perturbed structures when the target is the original clean operator. This is a coefficient-error mechanism consistent with the structural-noise interpretation. The AUC curves in Fig.~\ref{fig:Noise} quantify structural discrimination, whereas the expression above describes mean-square coefficient error. Dimension factors are suppressed here at fixed $E$; Sec.~S3 of the Supplementary Material states the distinction and assumptions.

\subsection{Benchmark comparisons, empirical topologies, and scaling}

Publicly available graph topologies provide useful structural stress tests for the reconstruction framework, since topology shapes collective response and synchronization sensitivity\cite{arenas2008synchronization,luo2026spectral}. In this setting, the empirical graphs are used only as fixed ground-truth interaction structures, while the steady-state observations are generated synthetically under controlled dynamics. We organize this part in three steps: comparison of reconstruction accuracy, comparison of dominant computational costs, and synthetic validation on empirical-topology suites.

\begin{figure}[t]
\centering
\includegraphics[width=0.9\linewidth]{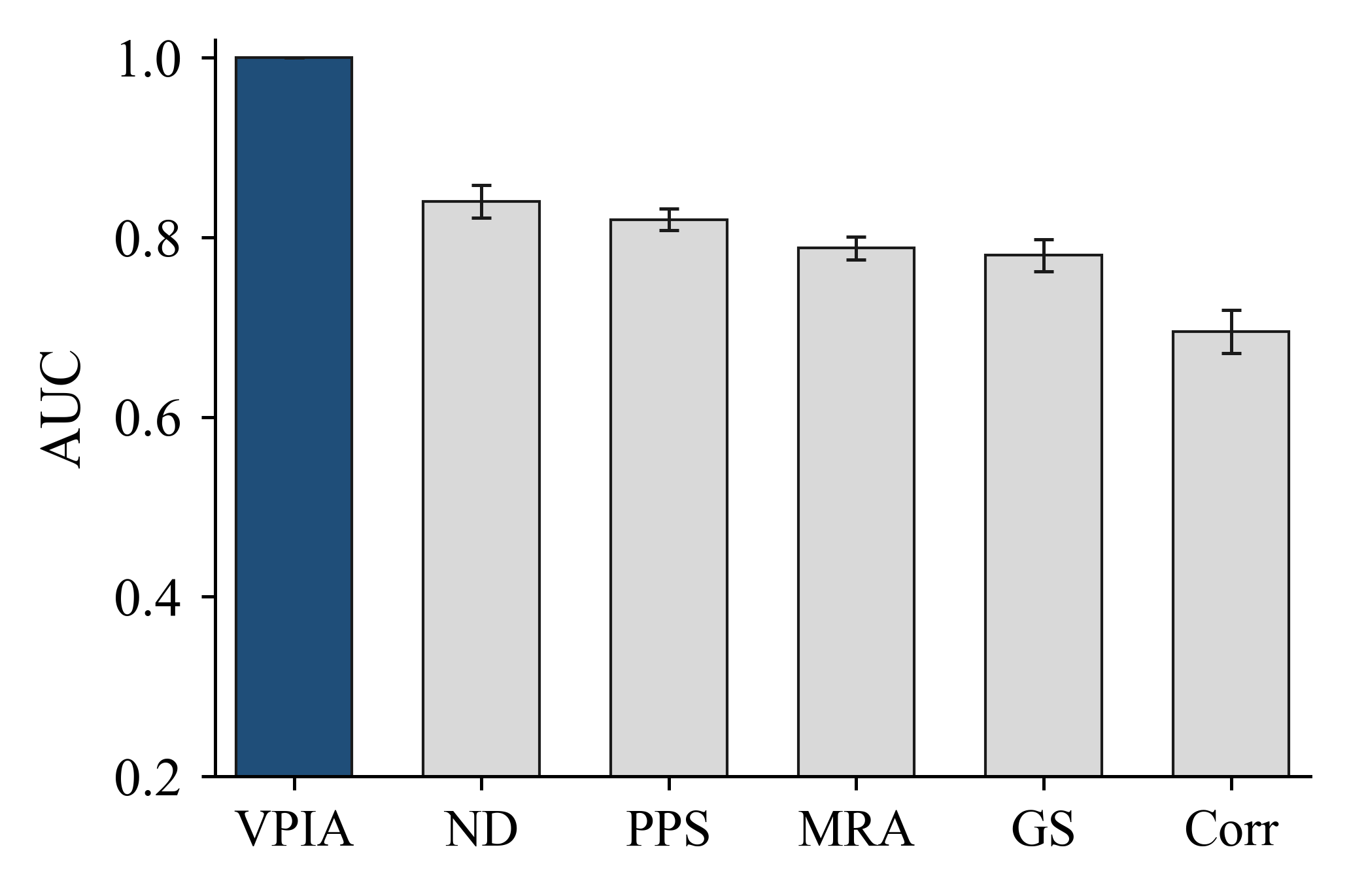}
\caption{\textbf{Accuracy comparison on an ER benchmark.}
Mean AUC for VPIA, ND, PPS, MRA, GS, and Corr at $N=50$. Bars and error bars show means and standard deviations over 20 realizations.}
\label{fig:realnet_baseline}
\end{figure}

\paragraph*{Accuracy comparison.}
We compare VPIA with network deconvolution (ND)\cite{feizi2013network}, partial phase synchronization (PPS)\cite{schelter2006partial}, modular response analysis (MRA)\cite{kholodenko2002mra}, global silencing (GS)\cite{barzel2013network}, and correlation scoring (Corr). VPIA attains the highest reported mean AUC in Fig.~\ref{fig:realnet_baseline}. These methods exploit different association, synchronization, response, and equilibrium information. Isolating the effect of parameterization requires a comparison with shared governing equations and controls.

A separate comparison over 100 independently seeded networks supplies VPIA, least squares (LS), and box-constrained least squares (CLS) with identical equilibrium equations and candidate edges. All three recover every tested binary support. At the stated settings, VPIA generally requires less inference time; Sec.~S6 of the Supplementary Material reports the seed distributions and analyzes the difference using convergence work, matrix conditioning, and coordinate-dependent stopping. The timing ordering depends on the solver settings and conditioning of the tested problems.

\paragraph*{Complexity comparison.}
Let $E_c$ be the number of candidate coefficients and $k_c$ the mean candidate degree. Pairwise sampled residual evaluation costs $O(Ms k_c)$. A dense update over all candidate parameters adds $O(E_c)$ work and optimizer storage; exact sparse accumulation requires an explicit sparse implementation. The present all-pairs model has $E_c=O(N^2)$ and $k_c=O(N)$ even when the planted graph is sparse. With $M=\Theta(N)$ and $s=\Theta(N)$, the residual-evaluation cost is cubic per iteration. Sampling a fixed residual budget describes a different computational regime, whose required iteration count must also be measured.

\begin{table}[t]
\centering\small
\caption{Dominant pairwise work for explicitly specified computational regimes. $T$ is the optimization iteration count. Candidate degree $k_c$ is distinct from true degree.}
\label{tab:complexity_comparison}
\begin{tabular}{lc}
\toprule
Operation & Work \\
\midrule
Sampled residuals & $O(Ms k_c)$ \\
Dense parameter update & $O(E_c)$ \\
Iterative inference & $O[T(Ms k_c+E_c)]$ \\
All-pairs, full residuals & $O(MN^2)$ per iteration \\
Dense $N\times N$ inversion & $O(N^3)$ per solve \\
\bottomrule
\end{tabular}
\end{table}

Table~\ref{tab:complexity_comparison} distinguishes per-update work from the cost of a complete solve. Sec.~S4 of the Supplementary Material additionally reports measured end-to-end time, including equilibrium generation and compatibility assembly, iteration counts, attained errors, and peak process memory as $N$ and $M$ vary. These finite-size measurements characterize the stated implementation and computational budget.

\paragraph*{Empirical-topology synthetic validation.}
We finally examine synthetic steady-state reconstruction on a broad suite of empirical topologies, summarized in Table~\ref{tab:Net}. The benchmark set ranges from small social networks with tens of nodes to a nearly thousand-node connectome-like graph, and it includes both undirected and directed structures; social-network topology can also organize hierarchical synchronization and propagation distortion\cite{luo2026hierarchical}. The published graphs serve as planted interaction operators from which heterogeneous steady states are generated under controlled dynamics.

\begin{table}[t]
\centering
\caption{Benchmark empirical graph topologies analyzed in this work.
Publicly available graph structures are used as fixed ground-truth interaction topologies for synthetic steady-state validation.
For each network, we report its type, number of nodes $N$, number of edges, and density $\rho$.}
\vspace{0.2cm}
\renewcommand{\arraystretch}{1.1}
\begin{tabular}{l c c c c}
\hline
Network & Type & Nodes & Edges & Density \\
\hline
Florentine Families & Undir & 16 & 20 & 0.1667 \\
Zachary Karate Club & Undir & 34 & 78 & 0.1390 \\
Wild Bird Social & Undir & 202 & 11733 & 0.5861 \\
\textit{C.~elegans} Neural & Dir & 297 & 2359 & 0.0259 \\
Power & Undir & 494 & 596 & 0.0049 \\
Retweet Weibo & Dir & 596 & 1415 & 0.0080 \\
Human Brain & Undir & 989 & 35730 & 0.0360 \\
\hline
\end{tabular}
\label{tab:Net}
\end{table}

\begin{figure*}
    \centering
    \includegraphics[width=\textwidth]{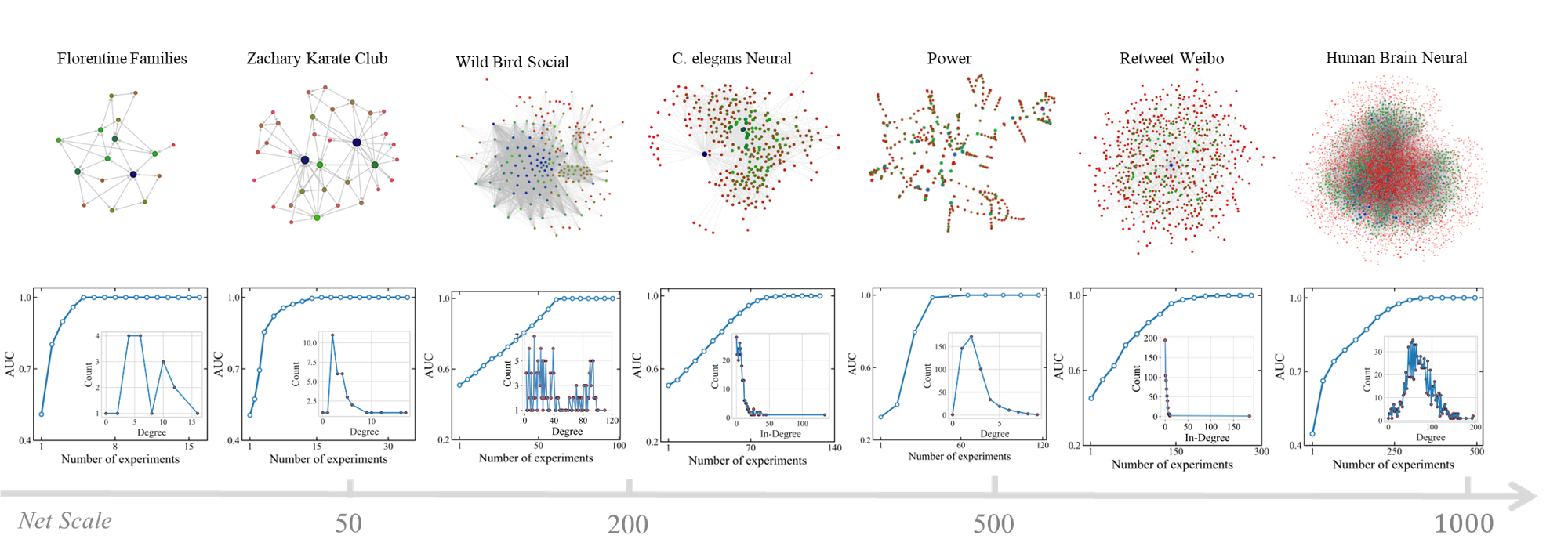}
    \caption{\textbf{Empirical-topology synthetic benchmarks.}
    Upper panels: representative benchmark graphs used as fixed planted interaction structures for synthetic steady-state tests, ranging from small social networks to directed communication networks and a large connectome-like graph.
    Lower panels: AUC as a function of the number of heterogeneous steady-state experiments for synthetic steady states generated on the same benchmark topologies.
    Insets show the corresponding degree or in-degree distributions.
    The empirical graphs serve as fixed planted interaction structures for synthetic steady states generated under known dynamics. The suite spans $N=16, 34, 202, 297, 494, 596,$ and $989$ across the seven panels, with $M$ varied along the horizontal axes and no added noise.}
    \label{fig:realnet}
\end{figure*}

Representative empirical-topology synthetic results are summarized in Fig.~\ref{fig:realnet}. The upper row shows that the benchmark suite spans compact social graphs, denser animal social networks, sparse infrastructure networks, directed communication networks, and a nearly thousand-node connectome-like topology. The lower row shows a common trend across these tested structures: AUC increases monotonically with the number of steady-state experiments and then approaches a topology-dependent saturation regime. Smaller graphs such as Florentine Families and Zachary Karate Club reach high discrimination after relatively few perturbations, whereas larger or more heterogeneous topologies such as Wild Bird Social, \textit{C.~elegans}, Retweet Weibo, and Human Brain require more experiments before entering the same high-AUC regime. The insets further indicate that this behavior persists across sharply peaked, heavy-tailed, and broad degree or in-degree distributions, suggesting that the reconstruction quality is not tied to a single structural ensemble within the tested synthetic protocols.

The accuracy comparison and empirical-topology suite support the practical use of equilibrium compatibility on the tested synthetic benchmarks. The retained topology demonstrations extend to nearly one thousand nodes, while the controlled timing tests address a smaller, explicitly reported range. Together, they assess structural recovery and computational performance under controlled synthetic dynamics.

\subsection{Extension to higher-order interactions}

The variational formulation extends naturally from pairwise adjacency matrices to tensor-valued interaction operators, retaining higher-order structure that can be invisible to the graph skeleton\cite{luo2026vulnerabilities}. This permits the same equilibrium-compatibility principle to be tested on selected higher-order systems in which triadic and higher-order couplings enter directly into the steady-state equations.

\begin{figure*}[t]
    \centering
    \includegraphics[width=0.8\textwidth]{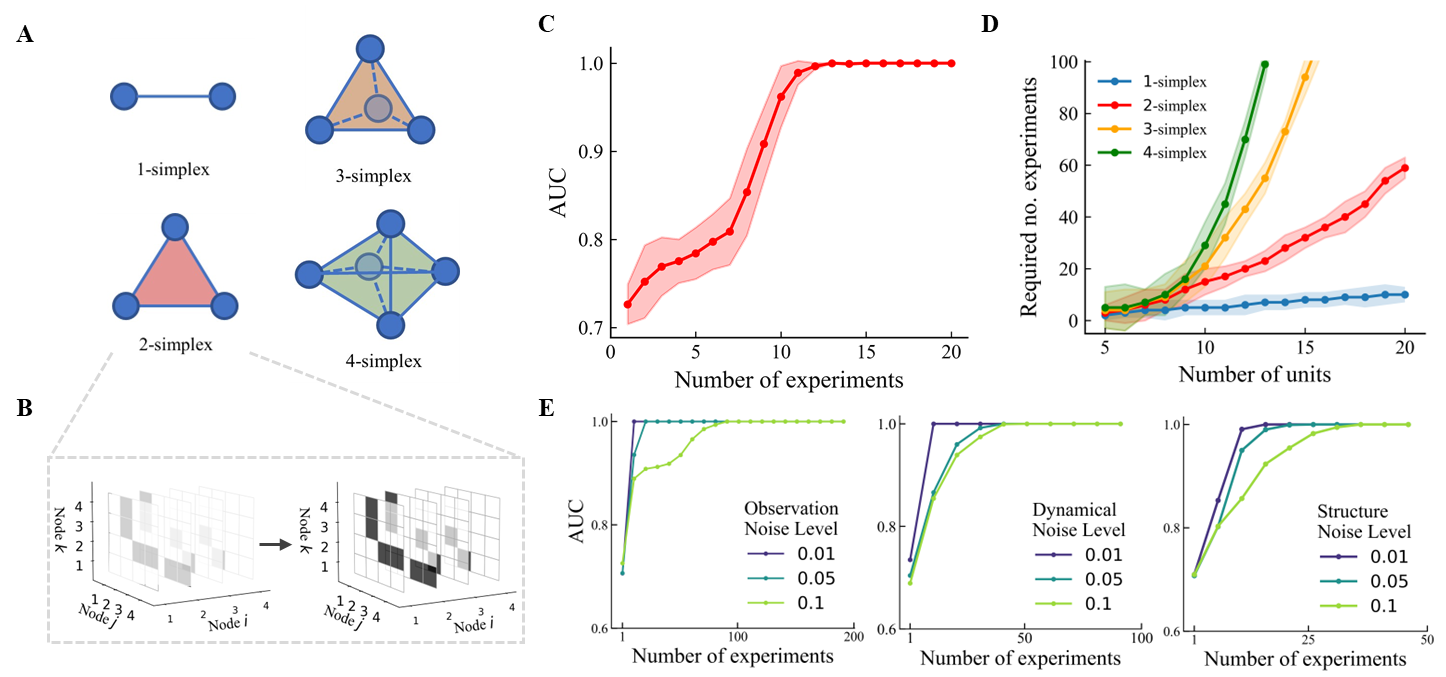}
    \caption{\textbf{Higher-order reconstruction.}
    (A) Illustrations of $1$-, $2$-, $3$-, and $4$-simplex interaction motifs, corresponding to two-, three-, four-, and five-vertex interaction structures.
    (B) Representative reconstruction of a $2$-simplex adjacency tensor.
    The left panel shows the planted higher-order tensor, and the right panel shows the tensor recovered by VPIA.
    (C) Reconstruction accuracy, measured by AUC, as a function of the number of steady-state experiments.
    (D) Number of steady-state experiments required for structural recovery across systems composed of $1$-, $2$-, $3$-, and $4$-simplex interactions.
    Higher-order simplices require more experiments because the number of unknown interaction coefficients grows combinatorially with simplex order.
    (E) Robustness under observation noise (left), dynamical noise (middle), and structural noise (right).
    Although larger noise amplitudes reduce the initial AUC, discrimination improves systematically as additional heterogeneous steady-state constraints are incorporated.
    These tests use dense higher-order systems of the sizes specified below. Panels (C)--(E) use networks with $N=10$, where all possible $2$-, $3$-, and $4$-simplex interactions are independently included with probability $p=0.5$. Panels C and E use $N=10$, panel D scans $N=5$ to $20$, $M$ is varied along the horizontal axes in panels C--E, and the noise amplitudes in panel E are $0.01$, $0.05$, and $0.1$.}
    \label{fig:higher_order}
\end{figure*}

Phase dynamics can provide information about simplex structure\cite{Luo2024}. A representative $2$-simplex reconstruction shows close agreement between the recovered tensor and the planted higher-order structure in Fig.~\ref{fig:higher_order}B. This indicates that higher-order interactions can also leave reconstructible signatures in the equilibrium constraints under sufficiently informative perturbations.

Reconstruction quality improves with the number of experiments, while the required sample count increases with simplex order in Fig.~\ref{fig:higher_order}C and Fig.~\ref{fig:higher_order}D. Section~II gives the corresponding affine-interaction identifiability bounds. The combinatorial growth of the coefficient dimension increases the data and computational requirements, motivating the moderate-size dense systems used here.

The higher-order extension also remains numerically stable under observation, dynamical, and structural perturbations in Fig.~\ref{fig:higher_order}E. As in the pairwise case, additional heterogeneous steady states improve reconstruction. At the same time, the rapidly increasing parameter dimension makes the higher-order setting intrinsically more demanding. These tests demonstrate reconstruction with tensor-valued interaction operators.

\section{Discussion}

The results clarify the setting in which heterogeneous steady states carry structural information about unknown interaction operators. When controlled perturbations generate diverse, nondegenerate stable equilibria and the steady-state equations depend informatively on the unknown coefficients, the accumulated compatibility constraints can isolate a unique operator within the affine-setting theory of Sec.~II. In this sense, the method converts repeated equilibrium measurements into an inverse problem whose information content is distributed across experiments rather than along a single trajectory.

\paragraph*{Identifiability and operating conditions.}
The analysis assumes known dynamics, controlled heterogeneous perturbations, and full node-level steady-state observations. Unique continuous-coefficient recovery in the affine setting requires a full-column-rank compatibility matrix. Near-synchronized equilibria can be uninformative, highly symmetric structures may reduce rank or worsen conditioning, and high-order parameter spaces grow combinatorially.

The numerical results suggest that heterogeneity improves reconstruction by generating linearly distinct equilibrium constraints. This point helps explain why reconstruction improves quickly once the perturbations produce sufficiently separated equilibria, and why filtering low-dispersion experiments is useful in practice. It also clarifies the interpretation of the empirical saturation curves: once the stacked constraints span the informative directions of the unknown operator, additional experiments mainly refine estimation error rather than change identifiability qualitatively.

From an inference standpoint, independent fluctuations can decrease with more experiments, while errors-in-variables bias and noise-induced mean shifts need not vanish. Quenched structural perturbations change the physical operator itself and can produce a floor relative to the clean structure. Computational cost is controlled by candidate interactions, the number of experiments, and convergence of the optimizer, rather than by true-network sparsity alone.

These considerations suggest several natural extensions. A first direction is perturbation design: because identifiability is controlled by the diversity of stable equilibria, one can ask how interventions should be chosen to maximize constraint informativeness. A second direction is reconstruction from partially observed equilibria, where hidden nodes or aggregate observables would require additional structural assumptions. A third direction is inference under controlled model misspecification, where the governing form is only approximately known. Within this scope, the results support heterogeneous steady states as informative observables for operator reconstruction under known dynamics.

\section{Conclusion}

VPIA provides a practical framework for steady-state network reconstruction in systems with known dynamical form, unknown interaction operators, controlled perturbations, and fully observed node-level equilibria. The central result is that, in this setting, reconstruction can be formulated as simultaneous enforcement of equilibrium compatibility across heterogeneous experiments, and in the affine-interaction class this leads to an exact continuous-coefficient rank criterion and necessary finite-sample counting bounds.

Across the numerical tests, this formulation supports accurate reconstruction on pairwise, directed, weighted, and selected higher-order examples within the tested settings. The numerical noise tests illustrate robustness within the tested regimes, while the corrected perturbation analysis separates variance reduction from possible bias. Residual sampling and diagonal preconditioning provide implementation options whose costs and convergence depend on the candidate set and data conditioning. Direct constrained solvers remain strong alternatives in the affine setting.

The results identify conditions under which heterogeneous steady states determine hidden interaction structure. When the dynamical form is known and controlled perturbations generate sufficiently heterogeneous equilibria, equilibrium observations can supply independent compatibility constraints that determine the hidden interaction operator. VPIA provides a practical variational implementation of this idea, while the affine theory clarifies the rank condition that ultimately controls identifiability.

\clearpage

\clearpage
\onecolumngrid
\section*{Supplementary Material}
\setcounter{section}{0}
\setcounter{subsection}{0}
\setcounter{subsubsection}{0}
\setcounter{equation}{0}
\setcounter{figure}{0}
\setcounter{table}{0}
\renewcommand{\thesection}{S\arabic{section}}
\renewcommand{\thesubsection}{S\arabic{section}.\arabic{subsection}}
\renewcommand{\thesubsubsection}{S\arabic{section}.\arabic{subsection}.\arabic{subsubsection}}
\renewcommand{\theequation}{S\arabic{section}.\arabic{equation}}
\renewcommand{\thefigure}{S\arabic{figure}}
\renewcommand{\thetable}{S\arabic{table}}
\renewcommand{\theHsection}{supp.\arabic{section}}
\renewcommand{\theHsubsection}{supp.\arabic{section}.\arabic{subsection}}
\renewcommand{\theHsubsubsection}{supp.\arabic{section}.\arabic{subsection}.\arabic{subsubsection}}
\renewcommand{\theHequation}{supp.\arabic{section}.\arabic{equation}}
\renewcommand{\theHfigure}{supp.\arabic{figure}}
\renewcommand{\theHtable}{supp.\arabic{table}}
\makeatletter
\@addtoreset{equation}{section}
\makeatother

\section{Kuramoto model in numerical experiments and dynamical models used to test in Fig. 2B}

This section summarizes all dynamical systems employed in the numerical
experiments in Fig.~2B in main text. For each model, we provide the standard form of the
governing equations and describe the external driving parameters that
are varied across steady-state experiments. These parameter perturbations
play the role of distinct experimental conditions $p^{(m)}$ in the main
text and generate the diverse steady states required for reconstruction.
All models are simulated until convergence to a stable equilibrium,
fixed point, or resting state, after which the final state is recorded
as the observed steady state $x^{*(m)}$. Only experiments yielding
stable heterogeneous equilibria after the filtering criterion of the
main text are retained.

\paragraph{Kuramoto model.}
\begin{equation}
\dot{\theta}_i = \omega_i
+ \sum_{j=1}^N A_{ij} \sin(\theta_j - \theta_i),
\label{eq:s1_kuramoto}
\end{equation}
where $\theta_i$ is the phase of oscillator $i$ and $\omega_i$ is its
intrinsic natural frequency. Across different steady-state experiments,
we perturb the entries of $\boldsymbol{\omega}$ to generate
heterogeneous fixed points.

\paragraph{Kuramoto--Sakaguchi model.}
\begin{equation}
\dot{\theta}_i = \omega_i
+ \sum_{j=1}^N A_{ij}
\sin(\theta_j - \theta_i - \alpha),
\end{equation}
where $\alpha$ is a phase-lag parameter, which is set to $0.1$. As in the Kuramoto case,
distinct experiments are obtained by modifying $\omega_i$.

\paragraph{Stuart--Landau oscillator.}
\begin{equation}
\dot{z}_i = (\lambda_i + i \omega_i - |z_i|^2)\,z_i
+ \sum_{j=1}^N A_{ij}(z_j - z_i),
\end{equation}
where $z_i \in \mathbb{C}$ is the complex amplitude. To generate
distinct steady states, we perturb the growth rate $\lambda_i$ or the intrinsic frequency $\omega_i$.  The parameters are set to $\lambda_i=1.0$.

\paragraph{Van der Pol oscillator.}
\begin{equation}
\begin{aligned}
\dot{x}_i &= y_i, \\
\dot{y}_i &= \mu_i(1 - x_i^2)y_i - x_i
+ \sum_{j=1}^N A_{ij}(x_j - x_i),
\end{aligned}
\end{equation}
where $\mu_i$ controls the nonlinearity. Across experiments, we adjust
$\mu_i$ or add small external biases to $x_i$.

\paragraph{FitzHugh--Nagumo model.}
\begin{equation}
\begin{aligned}
\dot{v}_i &= v_i - \frac{v_i^3}{3} - w_i + I_i
+ \sum_{j=1}^N A_{ij}(v_j - v_i), \\
\dot{w}_i &= \varepsilon (v_i + a - b w_i),
\end{aligned}
\end{equation}
where $v_i$ and $w_i$ are the activator and inhibitor variables.
Different experiments are produced by modifying the external input current $I_i$. The parameters are set to $a=0.7,~b=0.8,~\epsilon=0.08$.

\paragraph{R\"ossler system .}
\begin{equation}
\begin{aligned}
\dot{x}_i &= -\omega_i y_i - z_i 
+ \sum_{j=1}^N A_{ij}(x_j - x_i), \\
\dot{y}_i &= \omega_i x_i + a y_i, \\
\dot{z}_i &= b + z_i(x_i - c),
\end{aligned}
\end{equation}
where $(a,b,c)$ are selected in a stable fixed-point parameter range. 
We set
\[
a=0.05, \quad b=0.2, \quad c=4.0.
\]
Under these parameters, trajectories converge to a stable fixed point.
Distinct steady states are generated by small perturbations of $a$ 
or $\omega_i$ across experimental conditions.

\paragraph{Lorenz system .}
\begin{equation}
\begin{aligned}
\dot{x}_i &= \sigma (y_i - x_i) 
+ \sum_{j=1}^N A_{ij}(x_j - x_i), \\
\dot{y}_i &= x_i(\rho - z_i) - y_i, \\
\dot{z}_i &= x_i y_i - \beta z_i,
\end{aligned}
\end{equation}
where parameters are chosen below the chaotic threshold. 
We set
\[
\sigma=10.0, \quad \beta=8/3, \quad \rho \in [0.5,0.9].
\]
In this regime, all trajectories converge to a stable equilibrium.
Different steady states are generated by perturbing $\rho$.

\paragraph{Hindmarsh--Rose neuron model.}
\begin{equation}
\begin{aligned}
\dot{x}_i &= y_i - a x_i^3 + b x_i^2 - z_i + I_i
+ \sum_{j=1}^N A_{ij}(x_j - x_i), \\
\dot{y}_i &= c - d x_i^2 - y_i, \\
\dot{z}_i &= r(s(x_i - x_0) - z_i),
\end{aligned}
\end{equation}
with parameters
\[
a=1.0,~b=3.0,~c=1.0,~d=5.0,~r=0.01,~s=4.0,~x_0=-1.6.
\]
To remain in the resting-state (non-oscillatory) setting, 
external inputs are selected in the interval
\[
I_i \in [1.0,2.0].
\]
Distinct steady states are obtained by varying $I_i$ 
across experimental conditions.

\medskip
For all models, steady states are verified numerically by 
checking that the residual norms 
$\|\dot{\mathbf{x}}\|$, $\|\dot{\mathbf{y}}\|$, and 
$\|\dot{\mathbf{z}}\|$ fall below a prescribed tolerance 
after transient decay.

\section{Performance on Diverse Network Types, Directed and Weighted Structures}

This section provides additional numerical evidence across the tested network topologies and interaction types. We first show reconstruction performance on three representative network topologies---Barab\'asi--Albert (BA), Watts--Strogatz small-world (SW), and Erd\H{o}s--R\'enyi (ER) graphs---followed by evaluations on directed and weighted networks. These studies are numerical extensions of the pairwise core setting in the main text and are intended to document behavior across the tested ensembles rather than to enlarge the theoretical scope of Sec.~II. All experiments follow the same protocol as described in the main text.

\subsection{Reconstruction on Diverse Network Topologies}

In this subsection, we examine three different types of networks: Barab\'asi--Albert (BA), Watts--Strogatz small-world (SW), and Erd\H{o}s--R\'enyi (ER) networks. These topologies span heterogeneous structural regimes, from scale-free networks with heavy-tailed degree distributions (BA) to random networks with more homogeneous connectivity (ER).

\begin{figure}[t]
    \centering
    \includegraphics[width=0.75\linewidth]{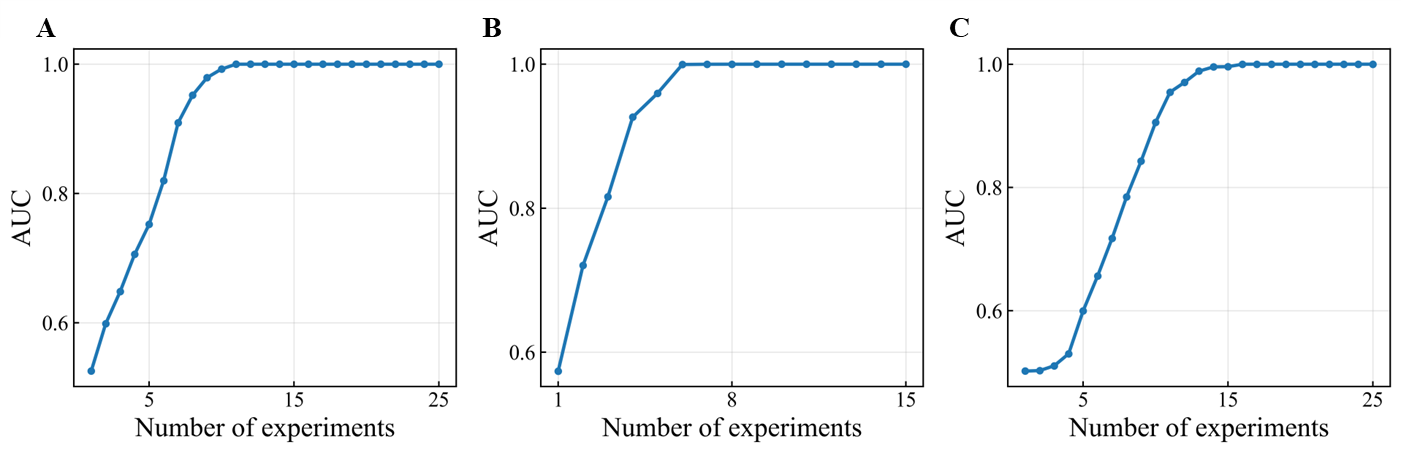}
    \caption[Reconstruction across different network topologies.]{
        \textbf{Reconstruction across different network topologies.}
        Illustration of true versus reconstructed interaction structures for three different network topologies:
        (a) Barab\'asi--Albert (BA) network with $N=100$, $m=2$,  
        (b) Watts--Strogatz small-world (SW) network with $N=100$, $k=4$, $\beta=0.1$, and  
        (c) Erd\H{o}s--R\'enyi (ER) network with $N=100$, $p=0.05$.  
        In all three topologies, VPIA recovers the main edge patterns, including hub nodes in BA networks, shortcut connections in SW networks, and random connectivity in ER networks. These examples are consistent with recovery across the tested structural ensembles.
    }
    \label{fig:topology_examples}
\end{figure}

As shown in Fig.~\ref{fig:topology_examples}(a)--(c), VPIA recovers the true network structure across these topologies. In the Barab\'asi--Albert (BA) network, it identifies the hub nodes characteristic of scale-free structure. In the Watts--Strogatz (SW) small-world network, it captures shortcut connections that bridge distant nodes. For the Erd\H{o}s--R\'enyi (ER) network, it recovers the random edge patterns. These results support consistent reconstruction across the tested ensembles and show that the same steady-state compatibility principle is not restricted to a single planted graph family.

\subsection{Reconstruction on Directed Networks}

Directed networks exhibit asymmetric couplings ($A_{ij}\neq A_{ji}$), resulting in distinct in-degree and out-degree statistics. These directional asymmetries are naturally encoded in the steady-state consistency equations of VPIA. In the affine counting argument of the main text, this ordered-pair parameterization doubles the number of independent pairwise coefficients relative to the undirected case and leads to the corresponding bound $M_{\mathrm{state,max}}=N$.

\begin{figure}[t]
    \centering
    \includegraphics[width=0.55\linewidth]{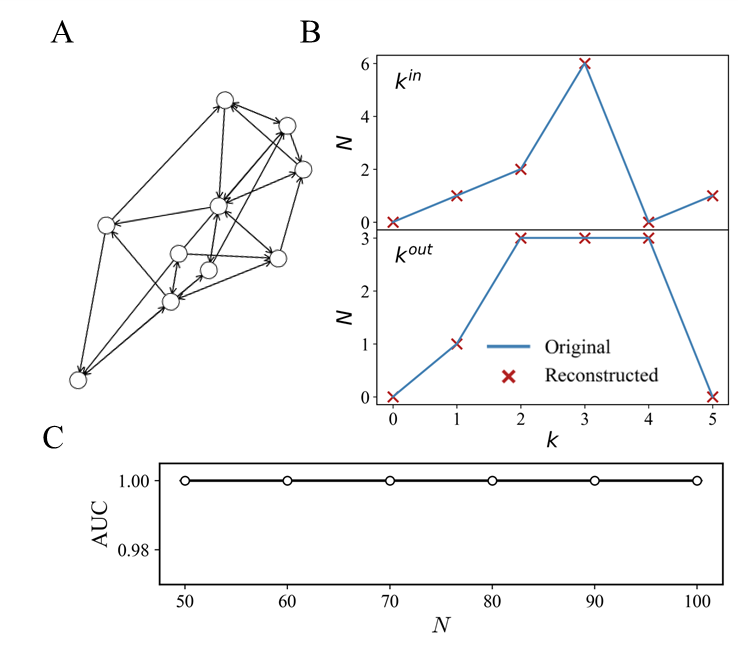}
    \caption[Directed network reconstruction.]{
        \textbf{Directed network reconstruction.}
        (a) Example directed network with $N=50$.  
        (b) Comparison between true and reconstructed in-/out-degree distributions, showing accurate capture of directed edge asymmetries.  
        (c) AUC for directed edge prediction over $N=50$--$100$ (averaged over 20 independent runs), consistently achieving high performance between $0.99$ and $1.00$.
    }
    \label{fig:directed_example}
\end{figure}

In this subsection, we examine directed-network reconstruction in the tested setting, where ordered pairs are inferred independently and the directional asymmetry must be recovered from the same heterogeneous steady-state constraints. The directed network shown in Fig.~\ref{fig:directed_example}(a) represents a small directed network with $N=50$ nodes. Fig.~\ref{fig:directed_example}(b) compares the true and reconstructed in- and out-degree distributions, showing that VPIA captures the correct asymmetry in the network. The AUC values in Fig.~\ref{fig:directed_example}(c), consistently between $0.99$ and $1.00$ across all tested sizes, support accurate reconstruction of asymmetric networks within the tested regime.

\subsection{Reconstruction on Weighted Networks}

Weighted networks extend the binary adjacency structure by assigning a real interaction strength $w_{ij}$ to each edge. To quantify continuous weight reconstruction accuracy, we compute the numerical error $\delta$ defined as:
\begin{equation}
\delta = \frac{1}{|E|}\sum_{(i,j)\in E} |\hat w_{ij} - w_{ij}|.
    \label{eq:delta_def}
\end{equation}
where $\hat{w}_{ij}$ denotes the inferred weight and $w_{ij}$ is the true weight.

\begin{figure}[t]
    \centering
    \includegraphics[width=0.55\linewidth]{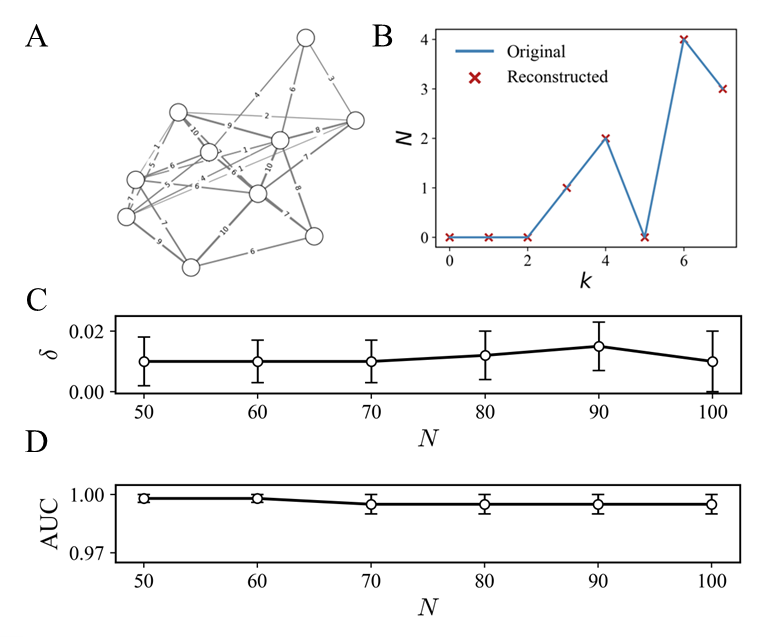}
    \caption[Weighted network reconstruction.]{
        \textbf{Weighted network reconstruction.}
        (a) Example weighted network with $N=50$ and integer weights.  
        (b) True and reconstructed degree distributions, showing that VPIA captures the degree structure accurately.  
        (c) Numerical weight error $\delta$ versus network size $N$, with errors consistently below $2\%$.  
        (d) AUC for binary support prediction, demonstrating high recovery accuracy with AUC values between $0.98$ and $1.00$.
    }
    \label{fig:weighted_example}
\end{figure}

Fig.~\ref{fig:weighted_example}(b) shows that the degree distributions of the true and reconstructed networks match well, indicating that the inferred weighted operator captures the dominant structural pattern. The numerical weight error in Fig.~\ref{fig:weighted_example}(c) remains in the $0$--$2\%$ range across all tested network sizes, supporting accurate recovery of continuous interaction amplitudes within the tested size range. In addition, Fig.~\ref{fig:weighted_example}(d) shows high AUC values ($0.98$--$1.00$) for binary support prediction, consistent with reliable support detection in the same tested setting.

As many empirical networks involve integer-valued edge weights, such as interaction counts or packet numbers, we additionally evaluate VPIA on networks with discrete edge weights. Fig.~\ref{fig:weighted_example}(a) illustrates that the same reconstruction objective can also be applied to discrete-weight test cases, supporting applicability within the tested setting.

\section{Robust identifiability under observational, dynamical, and structural noise}

This section analyzes how the affine-setting identifiability results of main text Sec.~II
are modified under three physically relevant sources of perturbations:
(i) measurement (observational) noise,
(ii) dynamical (process) noise, and
(iii) structural (quenched) noise in the interaction coefficients.
Although these noise mechanisms enter the reconstruction problem at
different stages of the physical process, their effects share a common
structure: at leading order, they induce zero-mean perturbations in the
linear constraints used for reconstruction, with variances controlled by
the corresponding noise intensities
$\sigma_{\mathrm{obs}},\sigma_{\mathrm{dyn}},\sigma_{\mathrm{str}}$.
For measurement and dynamical noise, fluctuations are independent across
experiments, and the reconstruction error decays as
$O(M^{-1/2})$. Structural noise, being quenched, produces an
irreducible floor of order $\sigma_{\mathrm{str}}$ with respect to the
clean structure, while the perturbed network itself remains recoverable
in the large-$M$ limit under the stated assumptions, and the clean
structure is recovered in expectation over the ensemble of structurally
perturbed networks. We now present the
three robustness results in detail.

\subsection{Robust identifiability under zero-mean noise}

We show that, in the presence of additive zero-mean noise of sufficiently small amplitude, the affine-setting reconstruction error can be made arbitrarily small by increasing the number of steady-state observations $M$. The argument is presented for the undirected 1-simplex case, where the interaction weights are collected into a vector $\mathbf a\in\mathbb{R}^E$ ($E$ is the number of edges). The same reasoning extends directly to higher-order interaction tensors within the affine-interaction setting of main text Sec.~II.

For each experimental condition $p^{(m)}$, the steady state $x^{(m)}\in\mathbb{R}^N$ satisfies a set of linear constraints on the unknown interaction parameters. As shown in main text Sec.~II, stacking all constraints from $M$ experiments yields the linear system
\begin{equation}
    H\mathbf a = \mathbf b ,
\end{equation}
where $H\in\mathbb{R}^{(N-1)M\times E}$ is the constraint matrix and $\mathbf b\in\mathbb{R}^{(N-1)M}$ the associated right-hand side. Under generic steady states, $H$ has full column rank and the unique noiseless solution is
\begin{equation}
    \mathbf a^\star = (H^\top H)^{-1} H^\top \mathbf b .
\end{equation}

We now incorporate observational noise. Let each steady-state measurement be corrupted by an additive perturbation of the form
\begin{equation}
    \tilde x^{(m)}_i=
    x^{(m)}_i + \sigma_{\mathrm{obs}}\,\mathrm{MAD}\,\xi_i ,
    \qquad 
    \xi_i\sim\mathcal N(0,1),
\end{equation}
where $\sigma_{\mathrm{obs}}>0$ is a dimensionless noise amplitude, $\mathrm{MAD}$ is the median absolute deviation of the steady states (providing a scale-normalized noise level), and $\xi_i$ are i.i.d.\ standard Gaussian variables. These perturbations propagate linearly to the right-hand side of the constraint system, producing the noisy equation
\begin{equation}
    \tilde{\mathbf b} = \mathbf b + \boldsymbol\eta ,
\end{equation}
with
\begin{equation}
    \mathbb{E}[\boldsymbol\eta]=\mathbf 0,
    \qquad
    \mathrm{Cov}(\boldsymbol\eta)
    = \sigma_{\mathrm{obs}}^2\,\mathrm{MAD}^2\, I .
\end{equation}

A natural reconstruction estimator is the least-squares solution
\begin{equation}
    \hat{\mathbf a}
    = 
    \arg\min_{\mathbf a}\|H\mathbf a - \tilde{\mathbf b}\|_2^2
    =
    (H^\top H)^{-1}H^\top \tilde{\mathbf b}.
\end{equation}
Substituting $\tilde{\mathbf b} = H\mathbf a^\star + \boldsymbol\eta$ yields
\begin{equation}
    \hat{\mathbf a}
    =
    \mathbf a^\star 
    +
    (H^\top H)^{-1} H^\top \boldsymbol\eta .
\end{equation}
Taking expectations gives
\begin{equation}
    \mathbb{E}[\hat{\mathbf a}] = \mathbf a^\star,
\end{equation}
showing that the estimator remains unbiased for all $\sigma_{\mathrm{obs}}>0$. Its covariance is
\begin{equation}
    \mathrm{Cov}(\hat{\mathbf a})
    =
    (H^\top H)^{-1}
    H^\top 
    \mathrm{Cov}(\boldsymbol\eta)
    H
    (H^\top H)^{-1}
    =
    \sigma_{\mathrm{obs}}^2\,\mathrm{MAD}^2\, (H^\top H)^{-1}.
\end{equation}

To extract the scaling with the number of experiments $M$, write the matrix $H$ as a vertical concatenation
\begin{equation}
    H 
    =
    \begin{pmatrix}
        H^{(1)}\\
        \vdots\\
        H^{(M)}
    \end{pmatrix},
    \qquad
    H^{(m)}\in\mathbb{R}^{(N-1)\times E}.
\end{equation}
Then
\begin{equation}
    H^\top H 
    = 
    \sum_{m=1}^M (H^{(m)})^\top H^{(m)}.
\end{equation}
Under generic experimental conditions, the matrices $(H^{(m)})^\top H^{(m)}$ are independent draws with finite second moments. By the law of large numbers,
\begin{equation}
    \frac{1}{M}H^\top H
    \xrightarrow[M\to\infty]{\text{a.s.}}
    Q,
\end{equation}
where the limiting matrix
\begin{equation}
    Q=\mathbb{E}\!\left[(H^{(m)})^\top H^{(m)}\right]
\end{equation}
is positive definite due to the identifiability condition. Hence for large $M$,
\begin{equation}
    H^\top H \approx MQ,
    \qquad
    (H^\top H)^{-1} \approx \frac{1}{M}Q^{-1}.
\end{equation}

Substituting this into the covariance expression yields the asymptotic scaling
\begin{equation}
    \mathrm{Cov}(\hat{\mathbf a})
    \approx 
    \sigma_{\mathrm{obs}}^2\,\mathrm{MAD}^2\, \frac{1}{M} Q^{-1}.
\end{equation}
Therefore the mean-squared reconstruction error satisfies
\begin{equation}
    \mathbb{E}\!\left[\|\hat{\mathbf a}-\mathbf a^\star\|_2^2\right]
    =
    \mathrm{Tr}\!\left(\mathrm{Cov}(\hat{\mathbf a})\right)
    =
    O\!\left(\frac{\sigma_{\mathrm{obs}}^2\,\mathrm{MAD}^2}{M}\right),
\end{equation}
and by Jensen's inequality,
\begin{equation}
    \mathbb{E}\!\left[\|\hat{\mathbf a}-\mathbf a^\star\|_2\right]
    =
    O\!\left(\frac{\sigma_{\mathrm{obs}}\,\mathrm{MAD}}{\sqrt{M}}\right).
\end{equation}

Thus, for any desired accuracy $\varepsilon>0$, choosing
\begin{equation}
    M \gtrsim \left(\frac{C\,\sigma_{\mathrm{obs}}\,\mathrm{MAD}}{\varepsilon}\right)^2 ,
\end{equation}
for a constant $C$ depending only on the limiting matrix $Q$, ensures that the 
reconstruction error is at most $\varepsilon$ with arbitrarily high probability as 
$M$ increases. The contributions of individual steady states accumulate 
coherently in $H^\top H$, while the noise averages out as $1/\sqrt{M}$. Hence, 
within the affine-setting assumptions and provided the noise amplitude does not 
destroy the conditioning of $H^\top H$, the reconstruction error can in principle 
be made arbitrarily small by increasing the number of steady-state experiments.

\subsection{Robust identifiability under dynamical noise}

We next analyze the dynamical-noise case within the same affine-setting framework. For sufficiently small noise intensity, the large-$M$ reconstruction error can again be made arbitrarily small, with the key difference that dynamical noise acts continuously during the system's evolution rather than only at measurement time. The argument parallels the measurement-noise case.

We consider a general networked dynamical system with state variables $x_i(t)$, interaction parameters $A$, and experiment-dependent external parameters $\lambda^{(m)}$. In the absence of dynamical noise, the evolution follows
\[
\dot x_i = F_i(x;A,\lambda),
\]
and experiment $m$ produces a steady-state outcome denoted by $x^{(m)}$.

To model dynamical fluctuations, we follow the discrete-time implementation used in numerical simulations. Let $\mathrm dt$ be the integration step, and let $\xi_i(t)\sim\mathcal N(0,1)$ be independent standard Gaussian variables. The simulation updates as
\begin{equation}
x_i(t+\mathrm dt)
=
x_i(t)
+
\mathrm dt\,F_i(x(t);A,\lambda)
+
\sigma_{\mathrm{dyn}}\sqrt{\mathrm dt}\,\xi_i(t),
\label{eq:s4_discrete_sde_final}
\end{equation}
where $\sigma_{\mathrm{dyn}}>0$ is the dynamical-noise amplitude. This is the Euler--Maruyama discretization of the stochastic differential equation
\begin{equation}
\mathrm d x_i(t)
=
F_i(x(t);A,\lambda)\,\mathrm dt
+
\sigma_{\mathrm{dyn}}\,\mathrm dW_i(t),
\label{eq:s4_continuous_sde_final}
\end{equation}
with $\{W_i(t)\}$ independent Wiener processes.

For each experiment $m$, Eq.~\eqref{eq:s4_discrete_sde_final} is simulated until the stochastic dynamics fluctuate around a stable equilibrium. Let $x^{(m)}(t)$ denote the resulting stochastic trajectory. The observed steady-state configuration used for reconstruction is defined as a time average over a window $[T_0,T_0+T]$,
\begin{equation}
\tilde x^{(m)}
=
\frac{1}{T}\!\int_{T_0}^{T_0+T} x^{(m)}(t)\,\mathrm dt.
\end{equation}
Standard results for time-averaged stochastic processes imply the decomposition
\begin{equation}
\tilde x^{(m)}
=
x^{(m)} + \delta x^{(m)},
\qquad
\mathbb{E}[\delta x^{(m)}]=0,
\qquad
\mathbb{E}\!\left[\delta x^{(m)}\delta x^{(m)\top}\right]
= O\!\left(\frac{\sigma_{\mathrm{dyn}}^2}{T}\right),
\label{eq:s4_time_avg_result_final}
\end{equation}
with variance reduced by $1/T$ due to time averaging.

We now examine how these perturbations enter the algebraic reconstruction constraints. In the noiseless setting, eliminating experiment-wise offsets yields a system of linear relations
\begin{equation}
H^{(m)}\mathbf a = \mathbf b^{(m)},
\label{eq:s4_constraint_deterministic_final}
\end{equation}
where $H^{(m)}$ and $\mathbf b^{(m)}$ depend smoothly on $x^{(m)}$ and $\lambda^{(m)}$, and $\mathbf a$ collects the interaction parameters. Substituting the noisy observations $\tilde x^{(m)} = x^{(m)} + \delta x^{(m)}$ produces perturbed coefficients
\[
\tilde H^{(m)} = H^{(m)} + \Delta H^{(m)},
\qquad
\tilde{\mathbf b}^{(m)} = \mathbf b^{(m)} + \Delta\mathbf b^{(m)}.
\]
A first-order Taylor expansion in $\delta x^{(m)}$ gives
\begin{equation}
\Delta H^{(m)} = \mathcal L^{(m)}[\delta x^{(m)}] + O(\|\delta x^{(m)}\|^2),
\qquad
\Delta\mathbf b^{(m)} = \mathbf r^{(m)}(\delta x^{(m)}) + O(\|\delta x^{(m)}\|^2),
\end{equation}
for appropriate linear operators $\mathcal L^{(m)}$ and vectors $\mathbf r^{(m)}$. Combining with Eq.~\eqref{eq:s4_time_avg_result_final},
\begin{equation}
\mathbb{E}[\Delta H^{(m)}] = O\!\left(\frac{\sigma_{\mathrm{dyn}}^2}{T}\right),
\qquad
\mathbb{E}[\Delta\mathbf b^{(m)}] = O\!\left(\frac{\sigma_{\mathrm{dyn}}^2}{T}\right),
\end{equation}
and similarly
\begin{equation}
\mathrm{Cov}(\mathrm{vec}(\Delta H^{(m)})) = O\!\left(\frac{\sigma_{\mathrm{dyn}}^2}{T}\right),
\qquad
\mathrm{Cov}(\Delta\mathbf b^{(m)}) = O\!\left(\frac{\sigma_{\mathrm{dyn}}^2}{T}\right).
\end{equation}

Stacking all $M$ experiments yields the noiseless global system
\[
H\mathbf a = \mathbf b,
\]
and the noisy system
\[
\tilde H\mathbf a = \tilde{\mathbf b},
\qquad
\tilde H = H + \Delta H,
\qquad
\tilde{\mathbf b} = \mathbf b + \boldsymbol\zeta,
\]
where $\Delta H$ and $\boldsymbol\zeta$ collect perturbations over all experiments. From the bounds above,
\[
\mathbb{E}[\Delta H],\ \mathbb{E}[\boldsymbol\zeta]
= O\!\left(\frac{\sigma_{\mathrm{dyn}}^2}{T}\right),
\qquad
\mathrm{Cov}(\mathrm{vec}(\Delta H)),\ \mathrm{Cov}(\boldsymbol\zeta)
= O\!\left(\frac{\sigma_{\mathrm{dyn}}^2}{T}\right).
\]

The least-squares estimator is
\[
\hat{\mathbf a} = (\tilde H^\top \tilde H)^{-1}\tilde H^\top \tilde{\mathbf b}.
\]
Expanding in $(\Delta H,\boldsymbol\zeta)$ and using the well-conditionedness of $H^\top H$,
\[
\hat{\mathbf a}
=
\mathbf a^\star
+
(H^\top H)^{-1}H^\top \boldsymbol\zeta
+
O(\|\Delta H\|\,\|\boldsymbol\zeta\| + \|\Delta H\|^2).
\]
Taking expectations gives
\[
\mathbb{E}[\hat{\mathbf a}]
=
\mathbf a^\star
+ O\!\left(\frac{\sigma_{\mathrm{dyn}}^2}{T}\right),
\]
so the estimator is asymptotically unbiased as $\sigma_{\mathrm{dyn}}^2/T\to 0$.

The covariance is dominated by
\[
\mathrm{Cov}(\hat{\mathbf a})
=
(H^\top H)^{-1} H^\top \mathrm{Cov}(\boldsymbol\zeta) H (H^\top H)^{-1}
+ \text{higher-order terms}.
\]
Writing $H$ as a stack of $M$ independent blocks, the law of large numbers yields
\[
H^\top H \approx M Q,
\qquad
(H^\top H)^{-1}\approx \frac{1}{M}Q^{-1},
\]
with $Q\succ 0$. Since each experiment contributes noise of variance $O(\sigma_{\mathrm{dyn}}^2/T)$,
\[
\mathrm{Cov}(\boldsymbol\zeta) = O\!\left(\frac{\sigma_{\mathrm{dyn}}^2}{T}\right)I,
\]
and thus
\[
\mathrm{Cov}(\hat{\mathbf a})
=
O\!\left(\frac{\sigma_{\mathrm{dyn}}^2}{T}\cdot\frac{1}{M}\right)
Q^{-1}
+ \text{higher-order terms}.
\]

This implies the mean error scalings
\[
\mathbb{E}\!\left[\|\hat{\mathbf a}-\mathbf a^\star\|_2^2\right]
= O\!\left(\frac{\sigma_{\mathrm{dyn}}^2}{T}\cdot\frac{1}{M}\right),
\qquad
\mathbb{E}\!\left[\|\hat{\mathbf a}-\mathbf a^\star\|_2\right]
= O\!\left(\frac{\sigma_{\mathrm{dyn}}}{\sqrt{T}}\cdot\frac{1}{\sqrt{M}}\right).
\]

Thus, by choosing the averaging window $T$ sufficiently large to suppress the bias term and increasing the number of experiments $M$ to reduce the $1/\sqrt{M}$ fluctuations, the affine-setting prediction is again that the reconstruction error can be made arbitrarily small. Under these assumptions, zero-mean dynamical noise preserves the same large-$M$ recovery picture as in the observation-noise case.

\subsection{Robust identifiability under structural noise}

We finally examine the effect of structural noise, in which every off-diagonal interaction coefficient is perturbed by an additive, zero-mean random fluctuation. Let $\mathbf a^\star\in\mathbb{R}^E$ denote the vector of true interaction parameters, where $E$ is the number of independent parameters (for example, the $E=N(N-1)/2$ entries of a symmetric adjacency matrix in the undirected 1-simplex case). Structural noise is modeled by perturbing each true interaction coefficient by an independent zero-mean random variable. The structurally perturbed interaction vector is therefore defined as
\begin{equation}
\tilde{\mathbf a}
=
\mathbf a^\star + \boldsymbol\delta,
\qquad
\mathbb{E}[\boldsymbol\delta]=\mathbf 0,
\qquad
\mathrm{Cov}(\boldsymbol\delta)=\Sigma_\delta,
\label{eq:s5_structural_noise_model_revised}
\end{equation}
where $\boldsymbol\delta\in\mathbb{R}^E$ is the structural-noise vector. In the simplest independent-noise model, each off-diagonal entry is perturbed with variance of order $\sigma_{\mathrm{str}}^2$, so that $\Sigma_\delta = O(\sigma_{\mathrm{str}}^2) I$. Once drawn, $\boldsymbol\delta$ remains fixed across all experiments, making the perturbed interaction vector $\tilde{\mathbf a}$ a quenched random parameter rather than a fluctuating one.

The physical system evolves according to the structurally perturbed interactions $\tilde{\mathbf a}$. Denote by $\tilde{x}^{(m)}$ the steady state of experiment $m$ obtained under the dynamics driven by $\tilde{\mathbf a}$. Neglecting measurement and dynamical noise for the moment, the steady states satisfy the exact algebraic relation
\begin{equation}
H(\tilde{x}^{(1)},\ldots,\tilde{x}^{(M)})\,\tilde{\mathbf a}
=
\mathbf b(\tilde{x}^{(1)},\ldots,\tilde{x}^{(M)}),
\label{eq:s5_exact_struct_system}
\end{equation}
where $H(\cdot)$ and $\mathbf b(\cdot)$ are constructed as described in main text Sec.~II. Under the identifiability conditions of main text Sec.~II, this system has the unique solution $\tilde{\mathbf a}$. Therefore, if the reconstruction procedure is applied to infinitely many noiseless steady states, it converges to the structurally perturbed interactions $\tilde{\mathbf a}$.

In practice, reconstruction is carried out using a finite number of experiments and using steady states that may also be corrupted by measurement noise or dynamical noise. As in the preceding noise subsections of this section, these effects can be modeled by an approximate system of linear constraints
\begin{equation}
H\mathbf a \approx \tilde{\mathbf b},
\end{equation}
where $H$ is the design matrix constructed from the observed steady states and $\tilde{\mathbf b}$ is the corresponding right-hand side. The least-squares estimator is
\begin{equation}
\hat{\mathbf a}
=
(H^\top H)^{-1}H^\top \tilde{\mathbf b}.
\end{equation}
This expression may be rewritten as
\begin{equation}
\hat{\mathbf a}
=
\tilde{\mathbf a}
+
\bigl[(H^\top H)^{-1}H^\top \tilde{\mathbf b} - \tilde{\mathbf a}\bigr],
\end{equation}
which separates the estimate into the structurally perturbed interactions $\tilde{\mathbf a}$ and a fluctuation term arising from finite-$M$ sampling and additional noise sources.

Under the same assumptions used in the derivations above (generic steady states, full-rank $H$, sufficiently small measurement/dynamical noise), the fluctuation term is unbiased around $\tilde{\mathbf a}$ and has variance that decreases as $O(1/M)$. Thus,
\begin{equation}
\mathbb{E}\!\left[\hat{\mathbf a}\,\big|\,\tilde{\mathbf a}\right]
=
\tilde{\mathbf a},
\qquad
\mathbb{E}\!\left[\|\hat{\mathbf a}-\tilde{\mathbf a}\|_2^2\,\big|\,\tilde{\mathbf a}\right]
=
O\!\left(\frac{1}{M}\right).
\label{eq:s5_conditional_fluctuation}
\end{equation}
This means that for any fixed structurally perturbed realization $\tilde{\mathbf a}$, the reconstruction error with respect to $\tilde{\mathbf a}$ vanishes as $M\to\infty$. The perturbed structure is recoverable from sufficiently many steady states under the stated assumptions.

To quantify the recovery error relative to the clean interaction vector $\mathbf a^\star$, we write
\begin{equation}
\hat{\mathbf a}-\mathbf a^\star
=
(\hat{\mathbf a}-\tilde{\mathbf a})
+
(\tilde{\mathbf a}-\mathbf a^\star)
=
(\hat{\mathbf a}-\tilde{\mathbf a}) + \boldsymbol\delta.
\end{equation}
Taking expectations over both the reconstruction fluctuations and the structural noise, and using $\mathbb{E}[\boldsymbol\delta]=0$, yields
\begin{equation}
\mathbb{E}[\hat{\mathbf a}]
=
\mathbb{E}\!\left[\mathbb{E}\!\left[\hat{\mathbf a}\,\big|\,\tilde{\mathbf a}\right]\right]
=
\mathbb{E}[\tilde{\mathbf a}]
=
\mathbf a^\star,
\end{equation}
so that the reconstruction is unbiased in expectation over the ensemble of structurally perturbed networks.

For the mean-squared error, we compute
\begin{align}
\mathbb{E}\!\left[\|\hat{\mathbf a}-\mathbf a^\star\|_2^2\right]
&=
\mathbb{E}\!\left[\|\hat{\mathbf a}-\tilde{\mathbf a}\|_2^2\right]
+
2\,\mathbb{E}\!\left[(\hat{\mathbf a}-\tilde{\mathbf a})^\top(\tilde{\mathbf a}-\mathbf a^\star)\right]
+
\mathbb{E}\!\left[\|\tilde{\mathbf a}-\mathbf a^\star\|_2^2\right].
\end{align}
The cross term vanishes because the reconstruction fluctuation is conditionally zero-mean by Eq.~\eqref{eq:s5_conditional_fluctuation}. Therefore,
\begin{equation}
\mathbb{E}\!\left[\|\hat{\mathbf a}-\mathbf a^\star\|_2^2\right]
=
\mathbb{E}\!\left[\|\hat{\mathbf a}-\tilde{\mathbf a}\|_2^2\right]
+
\mathbb{E}\!\left[\|\boldsymbol\delta\|_2^2\right].
\end{equation}
The first term decays as $O(1/M)$, and the second term is the structural-noise variance,
\begin{equation}
\mathbb{E}\!\left[\|\boldsymbol\delta\|_2^2\right]
=
\mathrm{Tr}(\Sigma_\delta)
=
O(\sigma_{\mathrm{str}}^2).
\end{equation}
We obtain the scaling law
\begin{equation}
\mathbb{E}\!\left[\|\hat{\mathbf a}-\mathbf a^\star\|_2^2\right]
=
O\!\left(\frac{1}{M}\right) + O(\sigma_{\mathrm{str}}^2),
\qquad
\mathbb{E}\!\left[\|\hat{\mathbf a}-\mathbf a^\star\|_2\right]
=
O\!\left(\frac{1}{\sqrt{M}}\right) + O(\sigma_{\mathrm{str}}).
\end{equation}
Thus the total error decomposes into a statistical term that vanishes with $M$ and an irreducible structural-noise term of order $\sigma_{\mathrm{str}}$, which does not diminish with additional steady-state experiments.

In summary, for any fixed structural noise strength $\sigma_{\mathrm{str}}>0$, the noisy interaction vector $\tilde{\mathbf a}$ is recoverable in the large-$M$ limit under the stated assumptions, while the deviation between $\tilde{\mathbf a}$ and the clean structure $\mathbf a^\star$ is intrinsically bounded by $O(\sigma_{\mathrm{str}})$. In an ensemble sense, the recovered interactions are unbiased estimates of the true interactions $\mathbf a^\star$ and concentrate around them with variance $O(\sigma_{\mathrm{str}}^2)$, whereas the additional reconstruction variance induced by finite $M$ decays as $O(1/M)$. Structural noise therefore establishes a fundamental accuracy floor of order $\sigma_{\mathrm{str}}$ but does not prevent recovery of each structurally perturbed network from its steady states.

\section{Complexity comparison and per-iteration scaling}

We analyze the computational scaling of VPIA using the notation of the main text.
Let $N$ denote the number of nodes, $M$ the number of observed steady states,
and $\rho$ the network density defined through
\[
\mathrm{nnz}(A)=\rho N^2.
\]
At each optimization step, VPIA evaluates a randomly sampled subset of
\emph{$r$ node--state residuals},
where $r$ denotes the number of sampled residual constraints per iteration.
This definition is consistent throughout the main text and supplementary material.
Unless noted otherwise, all complexity estimates in this section refer to a
single optimization step under residual sampling; they do not include the
iteration count required for convergence, which is problem dependent.

\subsection{Interpretation of the asymptotic comparison}

The formulas collected below compare the dominant algebraic operations required by each method under a common steady-state input model. They do not account for constant factors, stopping criteria, implementation details, or hardware-dependent acceleration. Accordingly, the complexity comparison is intended only to contrast leading-order computational structure and should not be interpreted as a universal wall-clock ordering across implementations.

For pairwise interactions, the residual-sampling cost reported for VPIA is the baseline reference case used in the main text. The same sampling logic extends formally to selected higher-order parameterizations, but the local residual-evaluation cost then grows with the number of active simplices touching a node. The pairwise estimates in Table~I should therefore be read as the primary comparison benchmark, while the higher-order demonstrations in the main text remain moderate-size proof-of-concept studies.

\subsection{Full-batch residual evaluation}

Each steady state provides $N$ algebraic constraints,
yielding a total of $MN$ node--state residuals.
For pairwise interactions of the form
\begin{equation}
F_i(x)=\sum_{j\in\mathcal N(i)} A_{ij}\Phi(x_j),
\end{equation}
the computational cost of evaluating a single residual scales
with the node degree $k_i$.
Denoting the average degree by $k=\rho N$,
a full evaluation of all constraints scales as
\begin{equation}
T_{\mathrm{full}}
= O\!\left(\sum_{m=1}^{M}\sum_{i=1}^{N} k_i\right)
= O(MNk)
= O(M\rho N^2).
\end{equation}

In absolutely sparse networks with bounded average degree
($\rho = O(1/N)$),
this reduces to $O(MN)$.
However, even in this regime,
the multiplicative dependence on $M$
renders full-batch optimization increasingly expensive
as the number of perturbations grows.

\subsection{Residual sampling and per-iteration scaling}

Instead of evaluating all $MN$ constraints,
VPIA samples only $r$ residuals per iteration.
Since each sampled residual requires at most $O(\rho N)$ operations,
the resulting per-iteration complexity is

\begin{equation}
T_{\mathrm{iter}} = O(r\rho N).
\end{equation}

This scaling eliminates the multiplicative dependence on $M$
and makes each iteration independent of the total number of
steady-state experiments.

In dense regimes ($\rho = O(1)$),
\[
T_{\mathrm{iter}} = O(rN).
\]
If $r = O(N)$, the per-iteration cost becomes $O(N^2)$,
remaining strictly sub-cubic.

In absolutely sparse regimes
($\mathrm{nnz}(A)=O(N)$, equivalently $\rho=O(1/N)$),
\[
T_{\mathrm{iter}} = O(r).
\]
Under the common sampling choice $r=O(N)$,
this yields linear scaling $O(N)$.

\subsection{Parameter dimension and gradient structure}

For pairwise networks,
the parameter dimension scales as
\[
\dim(\theta)=O(\rho N^2).
\]
In absolutely sparse networks this reduces to $O(N)$.

Each sampled residual depends only on the local neighborhood
of a single node.
Consequently, the corresponding gradient vector
contains at most $O(\rho N)$ nonzero entries.
Across one iteration, the total number of nonzero gradient components
is therefore $O(r\rho N)$,
and no dense $N\times N$ matrices are constructed.

\subsection{Natural-gradient update}

The natural-gradient update is written as
\begin{equation}
\Delta\theta=-\eta F^{-1}g,
\end{equation}
where $g$ denotes the sampled stochastic gradient and
\begin{equation}
F=\frac{1}{r}
\sum_{(i,m)\in\mathcal S}
\nabla_\theta R_i^{(m)}
\nabla_\theta R_i^{(m)\top}.
\end{equation}

In practice, VPIA employs diagonal or block-diagonal
approximations of $F$.
Since Fisher statistics are accumulated over the same
sampled residual subset,
the cost of forming and applying the preconditioner
matches the residual-evaluation order,
namely $O(r\rho N)$.

\subsection{Unified scaling and comparison}

Combining residual evaluation and parameter updates,
the overall per-iteration complexity of VPIA satisfies

\begin{equation}
T_{\mathrm{iter}} = O(r\rho N).
\end{equation}

In dense networks with $r=O(N)$,
this yields $O(N^2)$ scaling.
In absolutely sparse networks,
it reduces to linear $O(N)$ scaling.

These per-iteration scaling relations remain strictly sub-cubic
across both dense and sparse regimes,
in contrast to representative steady-state reconstruction
methods that rely on dense matrix inversion,
Jacobian regression,
or eigen-decomposition.

A detailed comparison with representative steady-state reconstruction
frameworks is summarized in Table~\ref{tab:supp_complexity_comparison}.
While Table~I in the main text reports the per-iteration asymptotic orders
for large-scale regimes,
Table~\ref{tab:supp_complexity_comparison} provides an explicit breakdown
of the dominant computational steps underlying each method.

Classical steady-state Jacobian regression (SJR) and global silencing (GS)
combine node-wise regression with dense matrix inversion,
resulting in mixed quadratic--cubic scaling
$\mathcal{O}(MN^2 + N^3)$.
Partial phase synchronization (PPS),
network deconvolution (ND),
and modular response analysis (MRA)
require eigen-decomposition or inversion of
dense $N\times N$ matrices,
leading to $\mathcal{O}(N^3)$ complexity.
Local regression from dynamics eliminates matrix inversion
but retains $\mathcal{O}(MN^2)$ scaling
due to node-wise regression across all steady states.

In contrast, VPIA avoids Jacobian estimation,
dense regression,
and covariance inversion.
Its per-iteration complexity is governed solely by
residual sampling and sparse gradient evaluation,
yielding $\mathcal{O}(r\rho N)$ scaling.
This per-iteration order remains strictly sub-cubic
in both dense and absolutely sparse regimes.

\begin{table}[t]
\centering
\caption{Computational complexity of representative network reconstruction methods.
$N$: number of nodes; $M$: number of steady states; 
$\rho$: network density; $r$: number of sampled node--state residuals per iteration.}
\label{tab:supp_complexity_comparison}
\begin{tabular}{p{4.2cm} p{6.5cm} p{3.2cm}}
\hline
Method & Main computational steps & Time complexity \\
\hline
\textbf{VPIA}
& Sample $r$ residuals; Fisher-preconditioned update.
& $\mathcal{O}(r\rho N)$ \\[4pt]

Steady-state Jacobian regression (SJR)
& $N$ high-dimensional regressions.
& $\mathcal{O}(MN^2 + N^3)$ \\[4pt]

Local regression from dynamics
& Node-wise nonlinear regression across $M$ states.
& $\mathcal{O}(MN^2)$ \\[4pt]

Partial Phase Synchronization (PPS)
& Construct and invert synchronization matrix.
& $\mathcal{O}(N^3)$ \\[4pt]

Network Deconvolution (ND)
& Eigen-decomposition of interaction matrix.
& $\mathcal{O}(N^3)$ \\[4pt]

Modular Response Analysis (MRA)
& Inversion of global response matrices.
& $\mathcal{O}(N^3)$ \\[4pt]

Global Silencing (GS)
& Covariance estimation and inverse filtering.
& $\mathcal{O}(MN^2 + N^3)$ \\[4pt]
\hline
\end{tabular}
\end{table}

\medskip

In practice, $r$ is chosen as a small fraction of $N$
(typically $0.03N$--$0.10N$),
providing low-variance stochastic gradient estimates
while preserving sub-cubic per-iteration cost.

\end{document}